\documentclass{article}

    \PassOptionsToPackage{sort,numbers}{natbib}

\usepackage[final]{neurips_2024}

\usepackage{graphicx} 
\usepackage[utf8]{inputenc} 
\usepackage[T1]{fontenc}    
\usepackage[colorlinks=true,breaklinks=true,citecolor=green]{hyperref}       
\usepackage{url}            
\usepackage{booktabs}       
\usepackage{amsfonts}       
\usepackage{nicefrac}       
\usepackage{microtype}      
\usepackage{xcolor}         
\usepackage{amsmath} 
\usepackage{algorithm,algorithmicx,algpseudocode}
\usepackage{multirow}

\makeatletter
\DeclareRobustCommand\onedot{\futurelet\@let@token\@onedot}
\def\@onedot{\ifx\@let@token.\else.\null\fi\xspace}

\makeatother

\title{Resfusion: Denoising Diffusion Probabilistic Models for Image Restoration Based on Prior Residual Noise}

\author{
    Zhenning Shi \textsuperscript{1} \qquad
    Haoshuai Zheng\textsuperscript{1}\qquad
    Chen Xu\textsuperscript{1}\qquad 
    Changsheng Dong\textsuperscript{1}\qquad
    Bin Pan\textsuperscript{2}\qquad\\
    \textbf{Xueshuo Xie\textsuperscript{3}\qquad
    Along He\textsuperscript{1}$^\ast$\qquad
    Tao Li\textsuperscript{1,3}\thanks{Corresponding authors: litao@nankai.edu.cn, healong2020@163.com}\qquad
    Huazhu Fu\textsuperscript{4}
    }
    \\
    \textsuperscript{1}College of Computer Science, Nankai University \\
    \textsuperscript{2}School of Statistics and Data Science, Nankai University \\
    \textsuperscript{3}Haihe Lab of ITAI\\
    \textsuperscript{4}Institute of High Performance Computing, A*STAR\\
}

\begin{document}
\maketitle

\begin{abstract}
Recently, research on denoising diffusion models has expanded its application to the field of image restoration. Traditional diffusion-based image restoration methods utilize degraded images as conditional input to effectively guide the reverse generation process, without modifying the original denoising diffusion process. However, since the degraded images already include low-frequency information, starting from Gaussian white noise will result in increased sampling steps. We propose Resfusion, a general framework that incorporates the residual term into the diffusion forward process, starting the reverse process directly from the noisy degraded images. The form of our inference process is consistent with the DDPM. We introduced a weighted residual noise, named \textbf{resnoise}, as the prediction target and explicitly provide the quantitative relationship between the residual term and the noise term in resnoise. By leveraging a smooth equivalence transformation, Resfusion determine the optimal acceleration step and maintains the integrity of existing noise schedules, unifying the training and inference processes. The experimental results demonstrate that Resfusion exhibits competitive performance on ISTD dataset, LOL dataset and Raindrop dataset with only \textbf{five} sampling steps. Furthermore, Resfusion can be easily applied to image generation and emerges with strong versatility. Our code and model are available at \href{https://github.com/nkicsl/Resfusion}{https://github.com/nkicsl/Resfusion}.
\end{abstract}

\section{Introduction}
\label{sec:intro}
Denoising diffusion models~\cite{ho2020denoising, song2020denoising} have emerged as powerful and effective conditional generative models, demonstrating remarkable success in synthesizing high-fidelity data for image generation. \citet{saharia2022image} proved that these generative processes can be applied to image restoration by feeding degraded images as conditional input into the score network. SNIPS~\cite{kawar2021snips} combines annealed Langevin dynamics and Newton's method to arrive at a posterior sampling algorithm, exploring the generative diffusion processes to solve the general linear inverse problems. Based on these, many diffusion-based models were adapted for downstream image restoration tasks~\cite{ozdenizci2023restoring, whang2022deblurring, wang2023lldiffusion, hou2024global, guo2023shadowdiffusion, he2023reti, he2024diffusion}. For traditional diffusion-based models, the reverse process begins with Gaussian white noise, considering only the degraded images as the conditional input. This results in an increased number of sampling steps. Image restoration tasks often focus on restoring and editing specific high-frequency details while preserving crucial low-frequency information, such as the image structure. The degraded images used as conditional input inherently contain the low-frequency information. Therefore, initiating the reverse process from Gaussian white noise for image restoration tasks appears unnecessary and inefficient.

Consequently, some works have proposed to generate clean images directly from degraded images or noisy degraded images. InDI~\cite{delbracio2023inversion} restores clean images through the reverse process of direct iteration to degraded images; DDRM~\cite{kawar2022denoising} reformulate the image restoration tasks as inverse problems when the mapping between clean and degraded images is available; IR-SDE~\cite{luo2023image} directly models the image degradation process using mean-reverting SDE (Stochastic Differential Equations); I$^2$SB~\cite{liu20232} constructs a Schrödinger bridge between clean and degraded data distributions; Resshift~\cite{yue2024resshift} shifts the residual term from degraded low-resolution images to high-resolution images, performing the recovery in the latent space. \citet{liu2023residual} introduced the Residual Denoising Diffusion Models (RDDM), generalizing the diffusion process of InDI and I$^2$SB. RDDM points out that co-learning the residual term and the noise term can effectively improve the model performance. However, RDDM has some limitations. Firstly, RDDM predicts the residual term and the noise term separately, without explicitly specifying their quantitative relationship. Secondly, due to its forward process adopting an accumulation strategy for the residual term and the noise term, its forward and reverse processes are inconsistent with the DDPM~\cite{ho2020denoising}, which results in poor generalization and interpretability. Thirdly, RDDM requires the design of a complex noise schedule, as utilizing existing noise schedules would result in performance loss. 
\begin{figure*}[tp]
	\includegraphics[width=\linewidth]{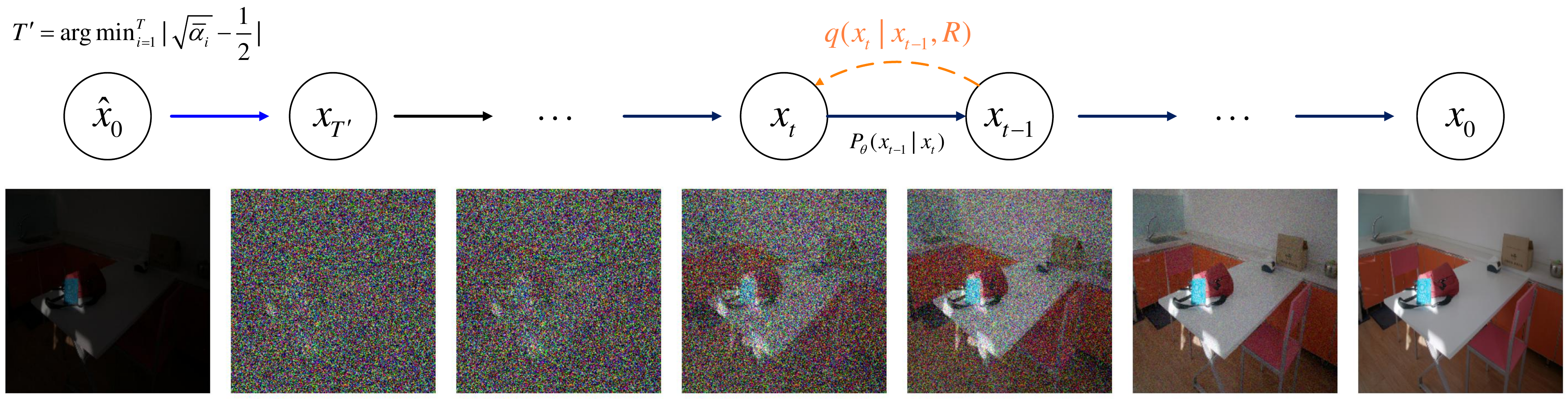}
	\caption{The proposed Resfusion is a general framework for image restoration and can be easily expand to image generation (setting $\hat{x}_{0}=0$). We introduce the residual term ($R = \hat{x}_{0}-x_{0}$) into the forward process, redefine $q(x_{t} | x_{t-1})$ to $q(x_{t} | x_{t-1}, R)$ (as shown by the $\color{orange}{orange}$ arrow), and name this diffusion process as resnoise diffusion. Through employing a novel technique called "smooth equivalence transformation", we can directly use the degraded image $\hat{x}_{0}$ to obtain $x_{T'}$ (as shown by the $\color{blue}{blue}$ arrow). We bridge the gap between the input image and ground truth, unifying the training and inference processes.}
	\label{fig: framework}
\end{figure*}

To solve the problems mentioned above, we propose \textbf{Resfusion}, a general framework for image restoration and can be easily expand to image generation. By introducing the residual term into the diffusion forward process, we bridge the gap between the input image and the ground truth, starting the reverse process directly from the noisy degraded images. We calculate the quantitative relationship between the residual term and the noise term, naming their weighted sum as the resnoise. Through the smooth equivalence transformation, we determine the optimal acceleration step and unify the training and inference processes. As a versatile methodology for image restoration, Resfusion does not require any physical prior knowledge. Resfusion allows for the direct use of existing noise schedules, and the image restoration process can be completed in just five sampling steps.

Our contributions can be summarized as follows:
\begin{itemize}
    \item First, by introducing the residual term into the diffusion forward process, our \textbf{resnoise-diffusion} process starts the reverse process directly from the noisy degraded images, closing the gap between the degraded input and the ground truth.
    \item Second, we explicitly provide the quantitative relationship between the residual term and the noise term in the loss function, and name the weighted residual noise as \textbf{resnoise}. Through transforming the learning of the noise term into the resnoise term, the form of our reverse inference process is consistent with the DDPM.
    \item Third, through the \textbf{smooth equivalence transformation} in resnoise-diffusion process, we determine the optimal acceleration step and unify the training and inference processes. The optimal acceleration step is non-trivial where the posterior probability distribution is equivalent to the prior probability distribution at this step. Moreover, we can directly use the existing noise schedule instead of redesigning the noise schedule. 
\end{itemize}

\section{Methodology}
\label{sec: methodology}
\subsection{Learning the resnoise}
\label{sec: learning the resnoise}
First, we denote the input degraded image and the ground truth as $\hat{x}_{0}$ and $x_{0}$. In order to extend the diffusion process of Denoising Diffusion Probabilistic Models (DDPM)~\cite{ho2020denoising} to image restoration, we define the residual term as Eq.~\eqref{eq: residual term}. 
\begin{equation}
	\label{eq: residual term}
	R = \hat{x}_{0}-x_{0}
\end{equation}
Then the forward process can be defined as Eq.~\eqref{eq: resfusion_q_sample_chain} and Eq.~\eqref{eq: resfusion_q_sample}. We provide a detailed explanation for why we introduce the residual term $R$ to Eq.~\eqref{eq: resfusion_q_sample} in this way in Sec.~\ref{sec: smooth equivalence transformation} and Figure~\ref{fig: working principle}. Consistent with the definition in DDPM, we use the notation $\beta_t = 1 - \alpha_t$ and $\overline{\alpha}_{t} =  \prod_{s=1}^{t}\alpha _s $.
\begin{equation}
	\label{eq: resfusion_q_sample_chain}
	q(x_{1:T} | x_{0}, R) = \prod_{t=1}^{T}q(x_{t}|x_{t-1}, R)
\end{equation}
\begin{equation}
	\label{eq: resfusion_q_sample}
	q(x_{t} | x_{t-1}, R) = N(x_{t} ; \sqrt{\alpha_{t}}x_{t-1} + (1-\sqrt{\alpha_{t}})R , (1-\alpha_{t})I)
\end{equation}
Then the redefined forward process can be formalized as Eq.~\eqref{eq: x_t_resnoise}, where $\epsilon$ represents the Gaussian white noise. 
\begin{equation}
	\label{eq: x_t_resnoise}
	{x}_{t}= \sqrt{\alpha_{t}} x_{t-1}+(1-\sqrt{\alpha_{t}})R+\sqrt{1-\alpha_{t}}\epsilon, \quad \epsilon\sim N(0,I)
\end{equation}
According to Eq.~\eqref{eq: x_t_resnoise}, ${x}_{t}$ can be reparameterized as Eq.~\eqref{eq: x_t_resnoise_reparameterized}.
\begin{equation}
	\label{eq: x_t_resnoise_reparameterized}
	{x}_{t}= \sqrt{\overline\alpha_{t}} x_{0}+(1-\sqrt{\overline{\alpha}_{t}})R+\sqrt{1-\overline\alpha_{t}}\epsilon, \quad \epsilon\sim N(0,I)
\end{equation}
We can easily incorporate this forward process into the vanilla DDPM. We introduce a residual noise, named as \textbf{resnoise} (symbolized as $res\epsilon$), to describe the gap between the current estimate $x_{t}$ and the ground truth $x_{0}$, and the term to be minimized can be formulated as Eq.~\eqref{eq: resnoise}. Detailed proof can be found in Appendix \ref{sec: Detailed proof}. 
\begin{equation}
	\label{eq: resnoise}
	res\epsilon = \epsilon + \frac{(1-\sqrt{\alpha_{t}})\sqrt{1-\overline{\alpha}_{t}}}{\beta_{t}}R, \quad
        \mathbb{E}_{x_{0},\epsilon,t}[||res\epsilon - res\epsilon_{\theta}(x_{t}, t)||^{2}]
\end{equation}
Through this process, we transform the learning of $\epsilon_{\theta }(x_{t},t)$ into $res\epsilon_{\theta}({x}_{t},t)$. $\epsilon_{\theta }(x_{t},t)$ represents the noise of the noisy ground truth, while $res\epsilon_{\theta}({x}_{t},t)$ represents the residual noise between the input degraded images and the ground truth. We name this process \textbf{resnoise-diffusion}. 

\subsection{Smooth equivalence transformation}
\label{sec: smooth equivalence transformation}
According to Eq.~\eqref{eq: x_t_resnoise_reparameterized} and Eq.~\eqref{eq: residual term}, we can derive Eq.~\eqref{eq: simple_x_t_resnoise}. It is worth mentioning that ${x}_T$ is uncomputable because the ground truth $x_{0}$ is unavailable in Eq.~\eqref{eq: simple_x_t_resnoise} during the reverse process, so we can not initialize ${x}_T$ directly. 
\begin{equation}
	\label{eq: simple_x_t_resnoise}
	{x}_{t}= (2\sqrt{\overline\alpha_{t}}-1) x_{0}+(1-\sqrt{\overline\alpha_{t}})\hat{x}_{0}+\sqrt{1-\overline\alpha_{t}}\epsilon, \epsilon\sim N(0,I)
\end{equation}
Fortunately, the weighted coefficient of $x_{0}$, which is $(2\sqrt{\overline\alpha_{t}}-1)$ in Eq.~\eqref{eq: simple_x_t_resnoise}, can be very close to zero. Since the input degraded image $\hat{x}_{0}$ is available, we can find a time step $T'$ where $x_{T'}$ is computable. 
When $T$ is sufficiently large, the variation of $\sqrt{\overline\alpha_{t}}(t \le T)$ with respect to time $t$ is smooth. We call this technique \textbf{smooth equivalence transformation}. Therefore, we can derive $T'$ as Eq.~\eqref{eq: T'} and obtain ${x}_{T'}$ in Eq.~\eqref{eq: P_x_T'} with a small bias. This bias can also be eliminated through the Truncated Schedule technique that we propose next.
\begin{equation}
	\label{eq: T'}
	T'=\arg\min_{i=1}^{T} |{\sqrt{\overline\alpha_{i}}}-\frac{1}{2}|
\end{equation}
\begin{equation}
	\label{eq: P_x_T'}
	x_{T'} \approx (1-\sqrt{\overline\alpha_{T'}})\hat{x}_{0}+\sqrt{1-\overline\alpha_{T'}}\epsilon
    \approx \sqrt{\overline\alpha_{T'}}\hat{x}_{0}+\sqrt{1-\overline\alpha_{T'}}\epsilon
\end{equation}
Thus we only need to minimize Eq.~\eqref{eq: resnoise} when $t \le T'$ since ${x}_{T'}$ is available, as shown in Eq.~\eqref{eq: accelerate diffusion}.
\begin{equation}
	\label{eq: accelerate diffusion}
	P_{\theta}(x_{0}) = \int_{{x}_{1}:{x}_{T'}} P_{data}({x}_{T'})
	\prod_{t=1}^{T'} P_{\theta}({x}_{t-1}|{x}_{t})d{x}_{1}:{x}_{T'}
\end{equation}
The resnoise-diffusion reverse process can be formulated as Eq.~\eqref{eq: accelerate} and Eq.~\eqref{eq: reverse process}. Consistent with the definition in DDPM, the $\Sigma_{\theta}$ is taken fixed as $\widetilde\beta_{t} = \frac{1-\overline{\alpha}_{t-1}}{1-\overline{\alpha}_{t}} \beta_t$.
\begin{equation}
	\label{eq: accelerate}
	P_{\theta }(x_{0:T'-1}|x_{T'})=\prod_{t=1}^{T'} P_{\theta}(x_{t-1}|x_{t})
\end{equation}
\begin{equation}
	\label{eq: reverse process}
	P_{\theta }(x_{t-1} | x_{t}) = N(x_{t-1} ; \mu_{\theta}(x_{t},t) , \Sigma_{\theta}(x_{t},t)), \quad P_{\theta }(x_{0} | x_{1}) = N(x_{0} ; \mu_{\theta}(x_{1},1))
\end{equation}
The mean $\mu_{\theta}(x_t, t)$ of resnoise-diffusion reverse process can be formalized as Eq.~\eqref{eq: mu_theta}, which is demonstrated in Appendix~\ref{sec: Detailed proof} from Eq.~\eqref{eq: widetilde_mu} to Eq.~\eqref{eq: Lt-1_simple}.
\begin{equation}
	\label{eq: mu_theta}
	\mu_{\theta}(x_t, t)=\frac{1}{\sqrt{\alpha_{t}}}(x_{t}-\frac{\beta_{t}}{\sqrt{1-\overline\alpha_{t}}}res\epsilon_{\theta }) 
\end{equation}
\begin{figure*}[t]
        \centering
		\includegraphics[width=0.8\linewidth]{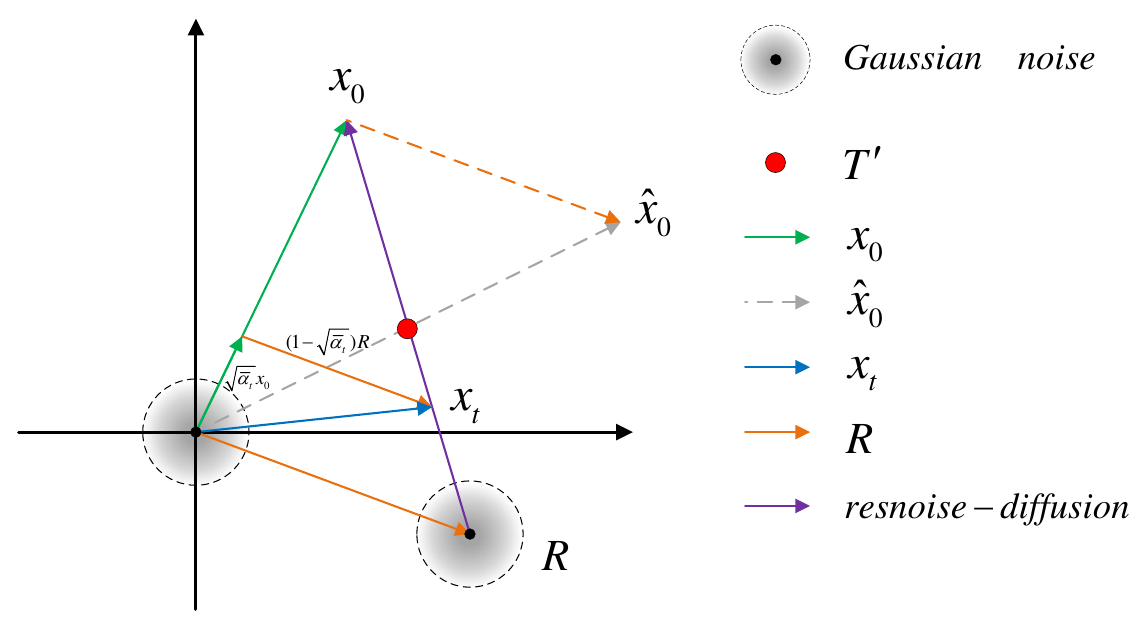}
    	\caption{The working principle of Resfusion. ${x}_{0}$ represents the distribution of the ground truth, while $\hat{x}_{0}$ represents the distribution of the degraded images. $\hat{x}_{0} - {x}_{0}$ represents the gap between them, defined as the residual term $R$ in Eq.~\eqref{eq: residual term}. Resfusion does not explicitly guide $\hat{x}_{0}$ to ${x}_{0}$. Instead, it implicitly learns the distribution of $R$ by doing resnoise-diffusion reverse process from $x_{t}$ to $x_{0}$. The resnoise-diffusion reverse process can be imagined as doing diffusion reverse process from $R+\epsilon$ to $x_{0}$ (as shown by the $\color{violet}{violet}$ arrow), guiding ${x}_{t}$ gradually towards ${x}_{0}$ along this direction. Following the principles of similar triangles, the coefficient of $R$ at step $t$ is computed as $1-\sqrt{\overline{\alpha}_{t}}$. At any step $t$ during the training process, ${x}_{t}$ can be calculated based on ${x}_{0}$ and $R$ through Eq.~\eqref{eq: x_t_resnoise}.}
		\label{fig: working principle}
\end{figure*}

Similar to previous work~\cite{liu2023residual, ozdenizci2023restoring}, our method enhances the diffusion model by incorporating a conditioning function. This function integrates latent representation from both the current estimate $x_{t}$ and the input degraded image $\hat{x}_{0}$.
Then Eq.~\eqref{eq: resnoise} can be modified to Eq.~\eqref{eq: simple_residual_term_with_It}.
\begin{equation}
	\begin{split}
	\label{eq: simple_residual_term_with_It}
   \mathbb{E}_{x_{0},\epsilon,t}[||res\epsilon - 
      res\epsilon_{\theta}((x_{t}, \hat{x}_{0}, t)||^{2}]
	\end{split}
\end{equation}
Just like the vanilla DDPM, Resfusion gradually fit the current estimate $x_{t}$ to the ground truth $x_{0}$, implicitly reducing the residual term $R$ between $\hat{x}_{0}$ and $x_{0}$ with the resnoise. 
When we define $\hat{x}_{t}$ as Eq.~\eqref{eq: hat_P_x_t} with the same Gaussian noise $\epsilon$ in $x_{t}$, $\hat{x}_{T'}$ can be seen as an intermediate result in an implicit DDPM process with the input degraded image $\hat{x}_{0}$ as the target distribution. We essentially quantitatively computed the accelerated step $T'$, on which step $x_{T'}$ is closed to $\hat{x}_{T'}$. When T is large enough, the approximate equal sign will become an equal sign. The implicit DDPM reverse process ($\epsilon$ to $\hat{x}_{0}$) is deterministic because $\hat{x}_{0}$ is available during the inference. The determination of step $T'$ corresponds to the point where the posterior probability distribution (resnoise-diffusion reverse process) becomes consistent with the prior probability distribution (implicit DDPM reverse process).
\begin{equation}
	\label{eq: hat_P_x_t}
	\hat{x}_{t}= \sqrt{\overline\alpha_{t}}\hat{x}_{0}+\sqrt{1-\overline\alpha_{t}}\epsilon, \quad \sqrt{\overline\alpha_{T'}} \approx 1-\sqrt{\overline\alpha_{T'}} \approx \frac{1}{2}, \quad x_{T'} \approx \hat{x}_{T'}
\end{equation}
As shown in Fig.~\ref{fig: working principle}, the resnoise-diffusion reverse process ($R+\epsilon$ to $x_{0}$) intersects with implicit DDPM reverse process ($\epsilon$ to $\hat{x}_{0}$), this intersection corresponds to step $T'$. The intersection of two diagonals of a parallelogram is the midpoint of them (corresponding to $0.5$ in Eq.~\eqref{eq: accelerate}), but due to the discrete nature of the diffusion process, the acceleration point actually falls on the point closest to the intersection, which is step $T'$ as Eq.~\eqref{eq: T'}. Meanwhile, since $x_{T'}$ is available, resnoise-diffusion steps after step $T'$ are not necessary according to Eq.~\eqref{eq: accelerate diffusion}. Therefore, Resfusion can directly start from step $T'$ for both inference and training process. Because the $\alpha$ coefficient of resnoise-diffusion is exactly the same as vanilla DDPM, Resfusion can directly use any existing noise schedule. In practical implementation, we utilized a technique called \textbf{Truncated Schedule} to control the offset between $x_{T'}$ and $\hat{x}_{T'}$ when $T$ is small. Further details can be found in Appendix \ref{sec: Truncated schedule}.

\section{Experiments}
\label{sec: experiments}
\begin{table*}[ht]%
\centering
\caption{Quantitative comparisons with other shadow removal methods. We report PSNR, SSIM~\cite{wang2004image} and MAE in the shadow region (S), the non-shadow region (NS) and all image (ALL). The best and second-best results are highlighted in \textbf{bold} and \underline{underlined}. ``$\uparrow$" (resp. ``$\downarrow$") means the larger (resp. smaller), the better. We use the symbol ``-" to indicate models or results that are unavailable.}%
\quad
\scalebox{0.7}{
\begin{tabular}{c|l|c|ccc| ccc| ccc}
        \toprule
        & \multicolumn{11}{c}{ISTD~\cite{wang2023ultra}}\\
        \cmidrule(l{.5em}r{.5em}){2-12}
        \multirow{2}{*}{} &\multirow{2}{*}{Method} &\multirow{2}{*}{Params} & \multicolumn{3}{c|}{Shadow Region (S)}  &
        \multicolumn{3}{c|}{Non-Shadow Region (NS)}  &
        \multicolumn{3}{c}{All Image (ALL)} \\
        & & & PSNR $\uparrow$ & SSIM $\uparrow$ & MAE $\downarrow$ & PSNR $\uparrow$ & SSIM $\uparrow$ & MAE $\downarrow$ & PSNR $\uparrow$ & SSIM $\uparrow$ & MAE $\downarrow$ \\
        \midrule
        \multirow{9}{*}{\rotatebox{90}{$256\times 256$}} &
        Input Image & -  & 22.40 & 0.936 & 32.11 & 27.32 & 0.976 & 6.83 & 20.56 & 0.893 & 10.97\\
        & ST-CGAN~\cite{wang2018stacked} & 31.8M & 33.74 & 0.981 & 9.99 & 29.51 & 0.958 & 6.05 & 27.44 & 0.929 & 6.65\\
        & DSC~\cite{hu2019direction} & 22.3M & 34.64 & 0.984 & 8.72 & 31.26 & 0.969 & 5.04 & 29.00 & 0.944 & 5.59\\
        & DHAN~\cite{cun2020towards} & 21.8M & 35.53 & 0.988 & 7.49 & 31.05 & 0.971 & 5.30 & 29.11 & 0.954 & 5.66\\
        & FusionNet~\cite{fu2021auto} & 186.5M & 34.71 & 0.975 & 7.91 & 28.61 & 0.880 & 5.51 & 27.19 & 0.945 & 5.88\\
        & UnfoldingNet~\cite{zhu2022efficient} & \underline{10.1M} & \underline{36.95} & 0.987 &8.29 & 31.54 & 0.978 & 4.55 & 29.85 & 0.960 & 5.09\\
        & DMTN~\cite{liu2023decoupled} & 22.8M & 35.83 & \underline{0.990} & 7.00 & 33.01 & \underline{0.979} & \underline{4.28} & 30.42 & \underline{0.965} & \underline{4.72}\\
        & RDDM (SM-Res-N)~\cite{liu2023residual} & 15.5M & 36.74 & 0.988 & \underline{6.67} & \underline{33.18} & \textbf{0.979} & \textbf{4.27} & \underline{30.91} & 0.962 & \textbf{4.67}\\
        & \textbf{Resfusion (ours)} & \textbf{7.7M} & \textbf{37.51} & \textbf{0.990} & \textbf{6.49} & \textbf{34.26} & 0.978 & 4.48 & \textbf{31.81} & \textbf{0.965} & 4.81\\
        \midrule
        \multirow{5}{*}{\rotatebox{90}{Original}} 
        & Input Image & - & 22.34 & 0.935 & 33.23 & 26.45 & \underline{0.947} & 7.25 & 20.33 & 0.874 & 11.35 \\
        & ARGAN~\cite{ding2019argan} & - & - & -& 9.21 & - & -&  6.27 & - & -&  6.63 \\
        & DHAN~\cite{cun2020towards}&  \underline{21.8M} & \underline{34.79} & \underline{0.983} & \underline{8.13} & \underline{29.54} & 0.941 & \underline{5.94} & \underline{27.88} & \underline{0.921} & 6.29 \\
        & CANet~\cite{chen2021canet} & 358.2M & - & - & 8.86 &- & - & 6.07 & -& -& \underline{6.15} \\   
        & \textbf{Resfusion (ours)} & \textbf{7.7M} & \textbf{36.45} & \textbf{0.985} & \textbf{7.08} & \textbf{32.08} & \textbf{0.950} & \textbf{5.02} & \textbf{30.09} & \textbf{0.932} & \textbf{5.34}\\
\bottomrule
\end{tabular}%
}
\label{tab: ISTD}%
\end{table*}
\begin{figure*}[ht]
		\centering
		\includegraphics[width=\linewidth]{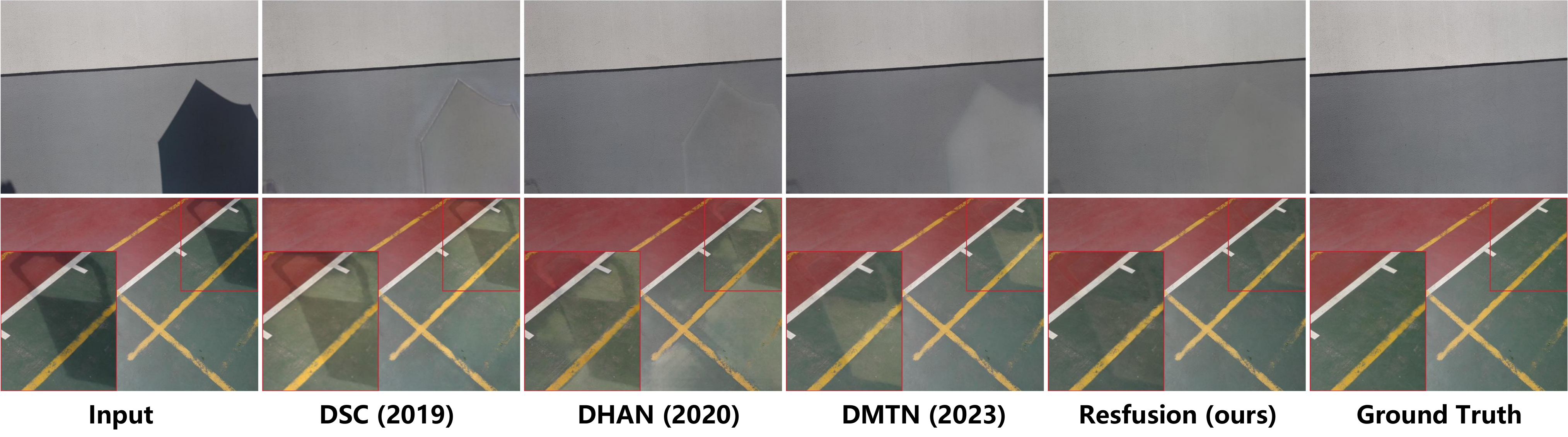}
    	\caption{Visual comparisons of the restored results by different shadow-removal methods on the ISTD dataset.}
		\label{fig: istd success}
\end{figure*}
To verify the performance of Resfusion, we conducted experiments on three image restoration tasks, including shadow removal, low-light enhancement and deraining. For fair comparisons, we use an U-net~\cite{ronneberger2015u} structure which is the \textbf{same} as RDDM as the backbone. We simply concatenate $x_t$ and $\hat{x}_{0}$ in the channel dimension and feed them into the network. For all the tasks, we only employ one U-net to predict resnoise. Furthermore, we simply utilize a truncated linear schedule~\cite{nichol2021improved} and perform only \textbf{five} sampling steps for all datasets. The experimental setting details are provided in the Appendix \ref{sec: experimental setting details}.

\textbf{ISTD dataset~\cite{wang2023ultra}} is a dataset designed for shadow removal, comprising 1870 sets of image triplets consisting of shadow image, shadow
mask, and shadow-free image. It consists of 1330 image triplets for training and 540 image triplets for quantitative evaluations.
We compare the proposed method with the popular shadow removal methods, i.e., ST-CGAN~\cite{wang2018stacked}, DSC~\cite{hu2019direction}, ARGAN~\cite{ding2019argan}, DHAN~\cite{cun2020towards}, FusionNet~\cite{fu2021auto}, CANet~\cite{chen2021canet}, UnfoldingNet~\cite{zhu2022efficient}, DMTN~\cite{liu2023decoupled}, and RDDM(SM-Res-N)~\cite{liu2023residual}. In order to ensure a fair comparison, we conducted experiments on two settings for the ISTD dataset, following the methods used in DMTN~\cite{liu2023decoupled} and DHAN~\cite{cun2020towards}: (1) The results are evaluated at a resolution of $256\times 256$ after being resized. (2) The original image resolutions ($640\times 480$) are maintained for evaluation.

\textbf{LOL dataset~\cite{wei2018deep}} comprises 500 pairs of images, consisting of both low-light and normal-light versions, which are further divided into 485 training pairs and 15 evaluation pairs. The low-light images contain noise produced during the photo capture process.
We compare the proposed method with the popular low-light enhancement methods, i.e., 
RetinexNet~\cite{wei2018deep}, KinD~\cite{zhang2019kindling}, KinD++~\cite{zhang2021beyond}, Zero-DCE~\cite{guo2020zero}, EnlightenGAN~\cite{jiang2021enlightengan}, Restormer~\cite{zamir2022restormer}, LLFormer~\cite{wang2023ultra}, RDDM (SM-Res-N)~\cite{liu2023residual}, and RDDM (SM-Res)~\cite{liu2023residual}. Some existing methods~\cite{wang2022low, hou2024global, zhou2023pyramid} calculate metrics by adjusting the overall brightness based on reference images (called as using GT-mean). However, this approach can introduce biases and potential unfairness. In accordance with LLFormer~\cite{wang2023ultra}, we compute metrics without utilizing any reference information. To ensure a fair comparison, we conducted experiments on two settings for the LOL dataset, following the methods employed in RDDM~\cite{liu2023residual} and LLFormer~\cite{wang2023ultra}: (1) The results are evaluated at a resolution of $256\times 256$ after being resized. PSNR and SSIM are evaluated based in YCbCr color space. (2) The original image resolutions ($600\times 400$) are maintained for evaluation. PSNR and SSIM are evaluated in RGB color space.
\begin{table}[t]%
\centering
\begin{minipage}{0.52\textwidth}
\centering
\caption{Quantitative comparisons with other low-light enhancement methods. We report PSNR, SSIM and LPIPS~\cite{zhang2018unreasonable}. The best and second-best results are highlighted in \textbf{bold} and \underline{underlined}. ``$\uparrow$" (resp. ``$\downarrow$") means the larger (resp. smaller), the better.}%
\scalebox{0.73}{
\begin{tabular}{l| c | ccc}
\toprule
\multicolumn{5}{c}{LOL~\cite{wei2018deep}}\\
\cmidrule{1-5}
Method & Params & PSNR $\uparrow$ & SSIM $\uparrow$ & LPIPS $\downarrow$\\
\midrule
\multicolumn{5}{c}{YCbCr space, $256\times 256$}\\
\midrule
Input Image & - & 9.30 & 0.377 & 0.513\\
RDDM (SM-Res-N)~\cite{liu2023residual} & 15.5M & 23.90 & 0.931 & -\\
RDDM (SM-Res)~\cite{liu2023residual} & \underline{7.7M} & \underline{25.39} & \underline{0.937} & \underline{0.116}\\
\textbf{Resfusion (ours)} & \textbf{7.7M} & \textbf{30.02} & \textbf{0.954} & \textbf{0.070}\\
\midrule
\multicolumn{5}{c}{RGB space, Original}\\
\midrule
Input Image & - & 7.77 & 0.191 & 0.560\\
RetinexNet~\cite{wei2018deep} & \underline{0.6M} & 16.77 & 0.560 & 0.474\\
KinD~\cite{zhang2019kindling} & 8.0M & 20.87 & 0.790 & 0.170\\
KinD++~\cite{zhang2021beyond} & 9.6M & 21.30 & \underline{0.820} & 0.160\\
Zero-DCE~\cite{guo2020zero} & \textbf{0.3M} & 14.86 & 0.562 & 0.335\\
EnlightenGAN~\cite{jiang2021enlightengan} & 8.6M & 17.48 & 0.652 & 0.322\\
Restormer~\cite{zamir2022restormer} & - & 22.37 & 0.816 & \underline{0.141}\\
LLFormer~\cite{wang2023ultra} & 24.6M & \underline{23.65} & 0.816 & 0.169\\
\textbf{Resfusion (ours)} & 7.7M & \textbf{24.63} & \textbf{0.860} & \textbf{0.107}\\
\bottomrule
\end{tabular}}%
\quad
\label{tab: LOL}%
\end{minipage}
\hspace{0.3cm}
\begin{minipage}{0.44\textwidth}
\centering
\caption{Quantitative comparisons with other deraining methods. We report PSNR and SSIM. The best and second-best results are highlighted in \textbf{bold} and \underline{underlined}. ``$\uparrow$" means the larger, the better.}%
\scalebox{0.73}{
\begin{tabular}{l | c |  cc}
\toprule
\multicolumn{4}{c}{Raindrop~\cite{qian2018attentive}}\\
\cmidrule{1-4}
Method & Params & PSNR $\uparrow$ & SSIM $\uparrow$ \\
\midrule
\multicolumn{4}{c}{YCbCr space, $256\times 256$}\\
\midrule
Input Image & - & 25.81 & 0.887\\
RDDM (SM-Res-N)~\cite{liu2023residual} & \underline{15.5M} & \underline{32.51} & \underline{0.956}\\
\textbf{Resfusion (ours)} & \textbf{7.7M} & \textbf{34.40} & \textbf{0.975}\\
\midrule
\multicolumn{4}{c}{YCbCr space, Original}\\
\midrule
Input Image & - & 25.40 & 0.882\\
pix2pix~\cite{isola2017image} & - & 28.02 & 0.855 \\
AttentiveGAN~\cite{qian2018attentive} & \textbf{6.2M} & 31.59 & 0.917 \\
DuRN~\cite{liu2019dual} & 10.2M & 31.24 & 0.926 \\
RaindropAttn~\cite{quan2019deep} & - & 31.44 & 0.926 \\
All-in-One~\cite{li2020all} & - & 31.12 & 0.927 \\
IDT~\cite{xiao2022image} & 16.4M & 31.87 & 0.931 \\
WeatherDiff$_{64}$~\cite{ozdenizci2023restoring} & 82.9M & 30.71 & 0.931 \\
RainDropDiff$_{128}$~\cite{ozdenizci2023restoring} & 109.7M & \underline{32.43} & \underline{0.933} \\
\textbf{Resfusion (ours)} & \underline{7.7M} & \textbf{32.61} & \textbf{0.938}\\
\bottomrule
\end{tabular}}%
\quad
\label{tab: Raindrop}%
\end{minipage}
\end{table}
\begin{figure*}[t]
		\centering
		\includegraphics[width=\linewidth]{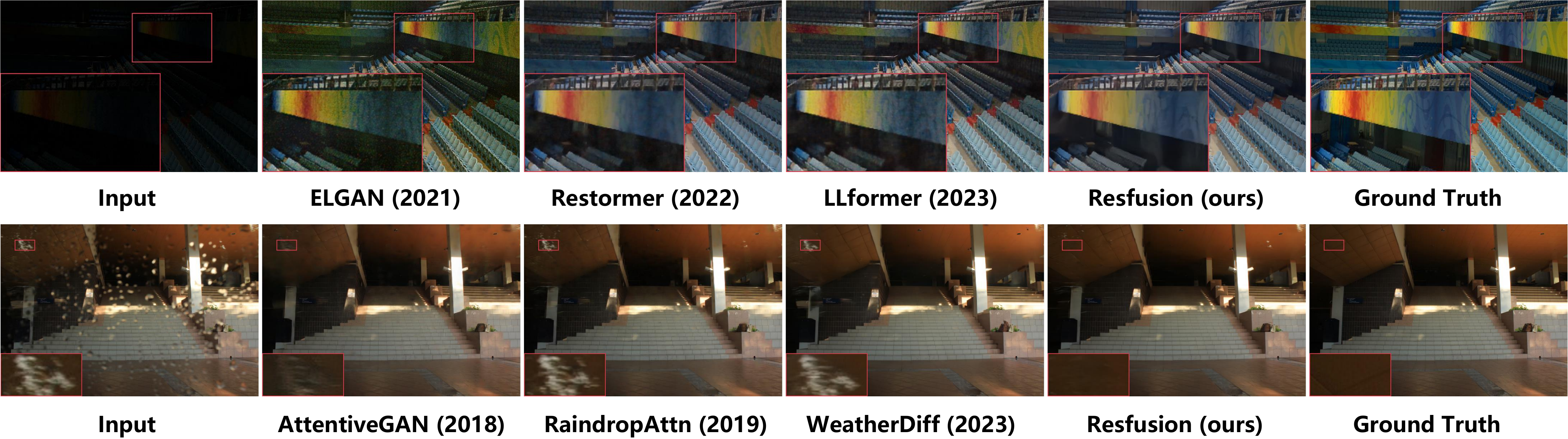}
    	\caption{Visual comparisons of the restored results by different image restoration methods on the LOL dataset and the Raindrop dataset.}
		\label{fig: lol+raindrop success}
\end{figure*}

\textbf{Raindrop dataset~\cite{qian2018attentive}} is a dataset designed for deraining, comprising 861 training image pairs for training, and 58 image pairs dedicated for quantitative evaluations, denoted in \cite{qian2018attentive} and \cite{ozdenizci2023restoring} as Raindrop-A.
We compare the proposed method with the popular deraining methods, i.e., pix2pix~\cite{isola2017image}, AttentiveGAN~\cite{qian2018attentive}, DuRN~\cite{liu2019dual}, RaindropAttn~\cite{quan2019deep}, All-in-One~\cite{li2020all}, IDT~\cite{xiao2022image}, WeatherDiff~\cite{ozdenizci2023restoring}, RainDropDiff~\cite{ozdenizci2023restoring}, and RDDM(SM-Res-N)\cite{liu2023residual}. We conduct experiments on two settings for the Raindrop dataset for fairness, following the methods employed in RDDM~\cite{liu2023residual} and WeatherDiff~\cite{ozdenizci2023restoring}: (1) The results are evaluated at a resolution of $256\times 256$ after being resized. (2) The original image resolutions are maintained for evaluation.

\textbf{Quantitative comparison}. As shown in the Tables \ref{tab: ISTD}, \ref{tab: LOL} \& \ref{tab: Raindrop}, We provide the quantitative evaluation results on ISTD dataset, LOL dataset and Raindrop dataset. Our methods clearly outperform all competing methods by significant margins in terms of PSNR, SSIM, MAE and LPIPS across all three datasets. The current experimental results demonstrate that Resfusion achieves highly competitive results under these conditions: (1) utilizing only one U-net to predict resnoise. (2) simply concatenating $x_t$ and $\hat{x}_{0}$ in the channel dimension. (3) employing a simple truncated linear schedule. (4) conducting only \textbf{five} sampling steps for all datasets. In contrast, alternative methods often rely on intricate network architectures, including multi-stage~\cite{fu2021auto, wang2023ultra, zhu2022efficient}, multi-scale~\cite{guo2023shadowformer}, multi-branch~\cite{cun2020towards} and prior knowledge of physics~\cite{wei2018deep, guo2020zero, quan2019deep, guo2023shadowdiffusion}, complex noise schedules~\cite{liu2023residual, yue2024resshift} and patch-overlapping strategy~\cite{ozdenizci2023restoring}.
For the ISTD dataset and Raindrop dataset, we only employ one U-net to predict resnoise, outperforming RDDM with two U-nets to predict the residual term and the noise separately in terms of PSNR and SSIM. Resfusion use \textbf{half} the number of parameters of RDDM and achieved better quantitative evaluation metrics.
For the LOL dataset, under the \textbf{same} parameters, Resfusion outperforms RDDM in terms of PSNR (+18\%) and LPIPS (-40\%) significantly. Furthermore, for all datasets, we employed a simple truncated linear schedule, while RDDM utilized a complex custom noise schedule. 

\section{Ablation Study}
\label{sec: ablation}
\subsection{The analysis of the residual term and the noise term}
\label{sec: The analysis of the residual term and the noise term}
\begin{figure*}[h]
    \centering
    \includegraphics[width=\linewidth]{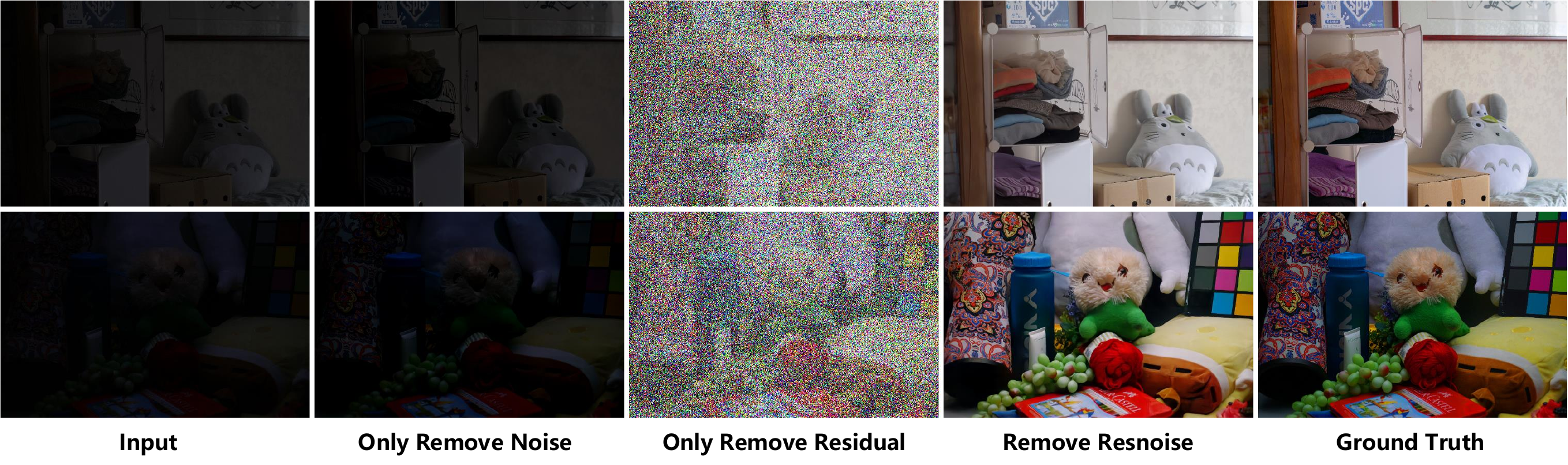}
    \caption{The analysis of the residual term and the noise term on the LOL dataset. Only removing noise will reconstruct the details of the degraded image without causing any semantic shift. Only removing residual can only accomplish the semantic shift (from low-light to normal-light) without reconstructing the details. Removing resnoise can achieve both the semantic shift and the detail reconstruction.}
    \label{fig: decoupling}
\end{figure*}
Resfusion and DDPM share the consist form of the reverse inference process. The DDPM reverse process restores the original image distribution by gradually removing the noise ($\epsilon_{\theta}$) from Gaussian white noise, while resnoise-diffusion reverse process restores clean image by gradually removing the resnoise ($res\epsilon_{\theta}$) from the noisy degraded images. Mathematically, the resnoise term ($res\epsilon_{\theta}$) and the noise term ($\epsilon_{\theta}$) only differ in the form of a weighted residual term ($\frac{(1-\sqrt{\alpha_{t}})\sqrt{1-\overline{\alpha}_{t}}}{\beta_{t}}R_{\theta}$). In other words, Resfusion subtracts an extra weighted residual term while removing the noise term during each step of the reverse process.

According to the Green’s theorem~\cite{riemann1867grundlagen}, when the neural network is sufficiently robust, the components of the resnoise should be path independent.  
Based on this belief, we trained two separate neural networks on the LOL dataset. One network predicts only $\epsilon_{\theta}$ and removes only the noise term during the resnoise-diffusion reverse process. The other network predicts only $\frac{(1-\sqrt{\alpha_{t}})\sqrt{1-\overline{\alpha}_{t}}}{\beta_{t}}R_{\theta}$ and removes only the weighted residual term during the resnoise-diffusion reverse process. 
As shown in Fig~\ref{fig: decoupling}, we qualitatively determine the functionalities of each component in the loss function of Resfusion. The weighted residual term are responsible for semantic shift, while the noise term handles detail reconstruction. Predicting resnoise $res\epsilon_{\theta}$ achieves both semantic shifts and detail reconstruction, bridging the gap between the input degraded image and the ground truth, enabling effective image restoration. 

\subsection{Equivalent representations of the loss function}
\label{sec: Equivalent representations of the loss function}
\begin{figure*}[h]
    \centering
    \includegraphics[width=\linewidth]{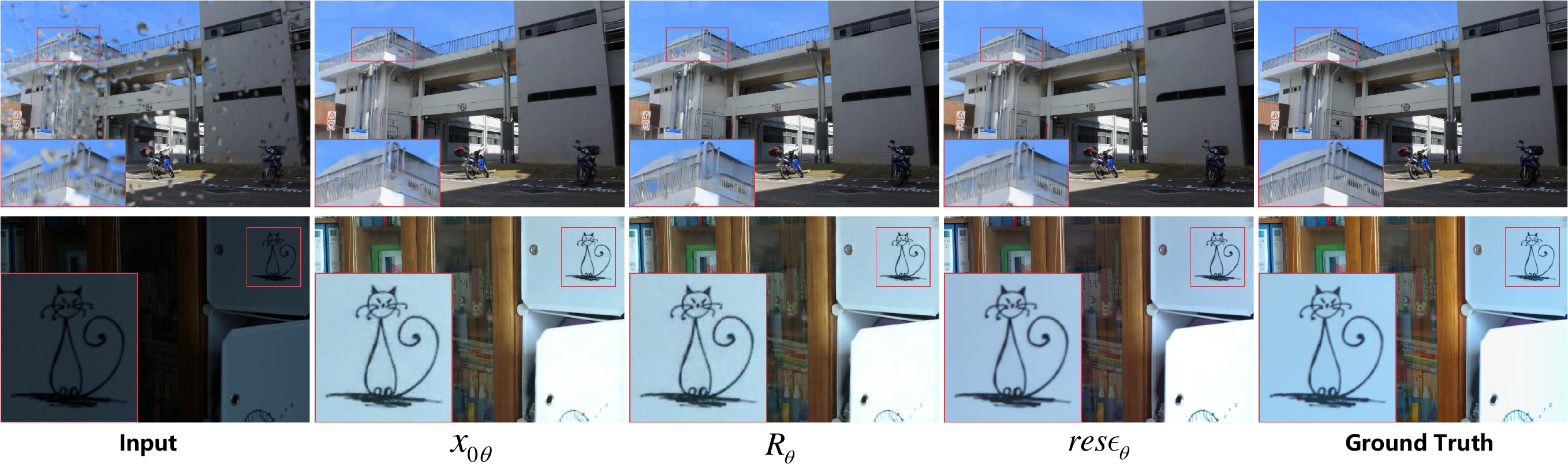}
    \caption{Visual comparisons of the restored results generated by other equivalent loss functions on Raindrop dataset and LOL dataset. Using $res\epsilon_{\theta}$ as the loss function allows for better restoration of details while completing semantic shifting.}
    \label{fig: equivalent loss functions}
\end{figure*}
Based on Eq.~\eqref{eq: residual term}, since $\hat{x}_{0}$ is available, once we acquire either $x_{0}$ or $R$, we can determine the other. Moreover, according to Eq.~\eqref{eq: q_sample_N}, since $x_{t}$ is obtainable and considering the equivalence between $x_{0}$ and $R$, we can also obtain the equivalent reverse process of resnoise-diffusion by predicting either $x_{0\theta}$ or $R_{\theta}$. Similar to DDPM, the fundamental purpose of Resfusion is also to predict $x_{0}$ (which is equivalent to predicting $R$). In contrast, DDPM reconstructs the distribution of the original image $x_{0}$ by incrementally reducing noise from the Gaussian white noise, whereas Resfusion skips the generation of low-frequency information by utilizing $\hat{x}_{0}$ for initialization. Therefore, in addition to the noise removal, Resfusion also involves removing a weighted residual term to accomplish the semantic shifting. It is worth mentioning that, unlike RDDM, predicting the noise term $\epsilon_{\theta}$ is \textbf{not} an equivalent loss function. Due to the truncated schedule we adopt, the starting point $x_{T'}$ will only consist of weighted $\hat{x}_{0}$ and $\epsilon$. As $\hat{x}_{0}$ is obtainable and serves as a conditional input to the neural network, directly predicting the noise term would lead to the neural network learning a simple pattern, resulting in training failure.

We compare the quantitative performance obtained by using different equivalent prediction targets as loss functions on ISTD dataset, LOL dataset and Raindrop dataset. We use the same backbone as described in Sec.~\ref{sec: experiments} and employ the same truncated linear schedule, performing five sampling steps for all datasets. The original image resolutions are maintained for evaluation. As shown in Table \ref{tab: equivalent loss functions}, predicting $res\epsilon_{\theta}$ outperformed predicting $x_{0\theta}$ and $R_{\theta}$ in terms of PSNR and SSIM on all three datasets. Predicting $res\epsilon_{\theta}$ resulted in better reconstruction of the fine details, as illustrated in Fig~\ref{fig: equivalent loss functions}.
\begin{table}[h]%
\centering
\caption{Quantitative comparisons with other equivalent loss functions on ISTD dataset, LOL dataset and Raindrop dataset. We report PSNR, SSIM, MAE and LPIPS. The best and second-best results are highlighted in \textbf{bold} and \underline{underlined}. ``$\uparrow$" (resp. ``$\downarrow$") means the larger (resp. smaller), the better.}%
\scalebox{0.85}{
\begin{tabular}{c | ccc | ccc | cc}
\toprule
\multirow{2}{*}{Prediction Targets} & & ISTD~\cite{wang2023ultra} & & & LOL~\cite{wei2018deep} & & \multicolumn{2}{c}{RainDrop~\cite{qian2018attentive}}\\
 & PSNR $\uparrow$ & SSIM $\uparrow$ & MAE $\downarrow$ & PSNR $\uparrow$ & SSIM $\uparrow$ & LPIPS $\downarrow$ & PSNR $\uparrow$ & SSIM $\uparrow$\\
\midrule
$x_{0\theta}$ & 29.67 & 0.927 & 5.35 & \underline{23.10} & \underline{0.813} & 0.150 & 32.51 & 0.935\\
$R_{\theta}$ & \underline{29.75} & \underline{0.930} & \textbf{5.26} & 22.87 & 0.807 & \underline{0.143} & \underline{32.57} & \underline{0.935}\\
$res\epsilon_{\theta}$ & \textbf{30.09} & \textbf{0.932} & \underline{5.34} & \textbf{24.63} & \textbf{0.860} & \textbf{0.107} & \textbf{32.61} & \textbf{0.938}\\
\bottomrule
\end{tabular}
}%
\label{tab: equivalent loss functions}
\end{table}

\section{Discussion}
\label{sec: discussion}
\begin{figure*}[ht]
    \centering
    \includegraphics[width=\linewidth]{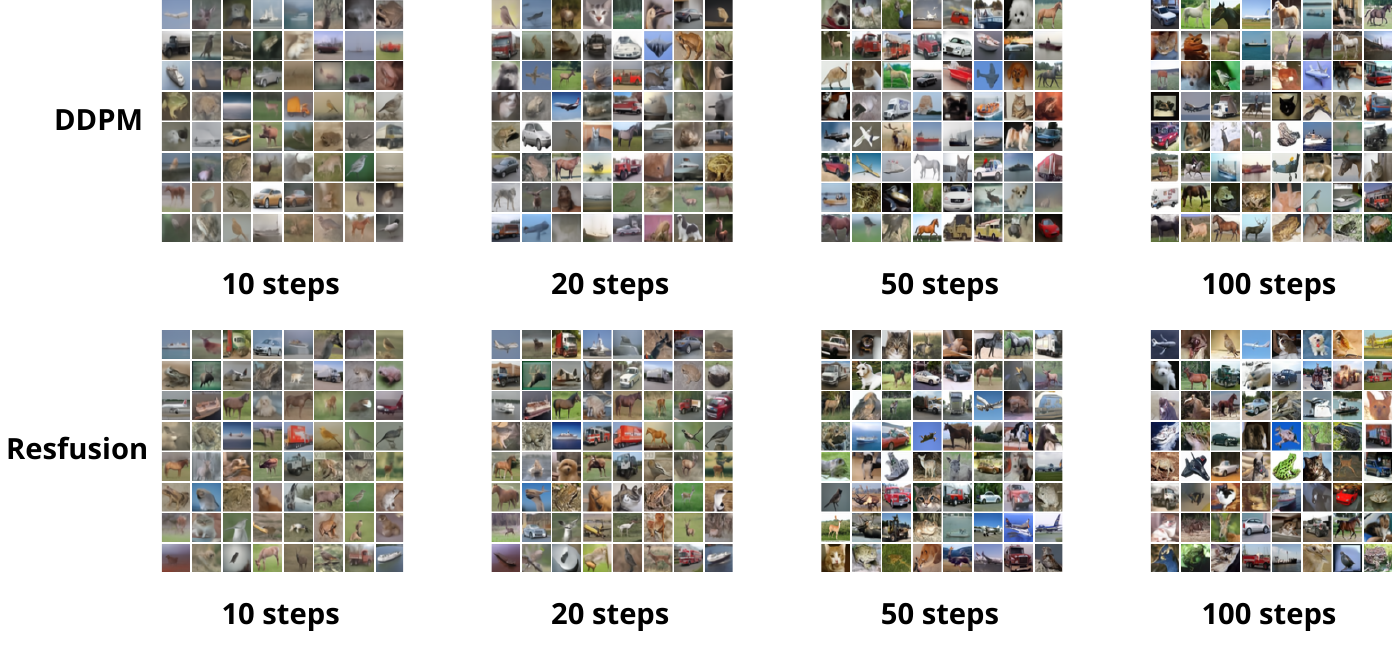}
    \caption{Visual comparisons between DDPM and Resfusion on the CIFAR10 ($32\times32$) dataset. We do not cherry-pick any results. With the same sampling steps, Resfusion outperforms DDPM in semantic generation and detail reconstruction.}
    \label{fig: generation}
\end{figure*}
Resfusion is not limited to image restoration. In fact, It is a versatile framework which can be applied to any general image generation domain. For image generation tasks, as it is impossible to obtain any additional information, we redefine $\hat{x}_{0}$ as a zero matrix. Thus we can obtain a new definition of the residual term $R=-x_{0}$ for image generation. Therefore, $res\epsilon$ is redefined as $res\epsilon=\epsilon - \frac{(1-\sqrt{\alpha_{t}})\sqrt{1-\overline{\alpha}_{t}}}{\beta_{t}}x_{0}$. Applied with the redefined residual term and the resnoise, Resfusion's forward and reverse process for image generation are completely consistent with the original resnoise-diffusion process for image retoration.

The redefined $res\epsilon$ further explains why Resfusion can complete the reverse inference process in fewer sampling steps than DDPM, under the same noise schedule. Due to consistency between the reverse processes, both of Resfusion and DDPM require the removal of the noise term. However, compared to DDPM, Resfusion additionally add a weighted $x_{0}$ instead of simply removing the noise. This is the key factor that allows Resfusion to diffuse faster.

We train DDPM, Resfusion, and DDIM~\cite{song2020denoising} on the CIFAR10 ($32\times32$) dataset~\cite{krizhevsky2009learning} with the same backbone. The experimental details are provided in the Appendix \ref{sec: experimental setting details}. A truncated linear schedule is used for Resfusion, and a linear schedule is used for DDPM and DDIM. We employ the Frechet Inception Distance (FID)~\cite{heusel2017gans} as the quantitative metric. As shown in Table \ref{tab: generation}, Resfusion significantly outperforms DDPM with the same sampling steps. At nearly half of the sampling steps, Resfusion achieves a similar FID with DDPM. Interestingly, when using a truncated linear schedule, the value of $T'/T$ is also closed to $0.5$, further validating Resfusion's accelerated sampling property. Consistent with DDPM, Resfusion also performs stochastic steps during the reverse process. Due to its stochastic nature, similar to DDPM, Resfusion's performance is lower than the deterministic DDIM.
\begin{table}[h]%
\centering
\caption{Quantitative comparisons with DDPM and DDIM on CIFAR10 ($32\times32$) dataset. We report FID under different sampling steps. ``$\downarrow$" means the smaller, the better.}%
\scalebox{0.85}{
\begin{tabular}{c | c | c | c}
\toprule
CIFAR10 (FID $\downarrow$) & DDPM & Resfusion(ours) & DDIM\\
\midrule
10 steps  & 43.11 & 28.81 & 18.37\\
20 steps  & 24.88 & 15.46 & 10.93\\
50 steps  & 14.02 & 7.96 & 7.39\\
100 steps & 9.79 & 6.31 & 6.21\\
\bottomrule
\end{tabular}
}%
\label{tab: generation}
\end{table}

\begin{figure*}[ht]
		\centering
		\includegraphics[width=\linewidth]{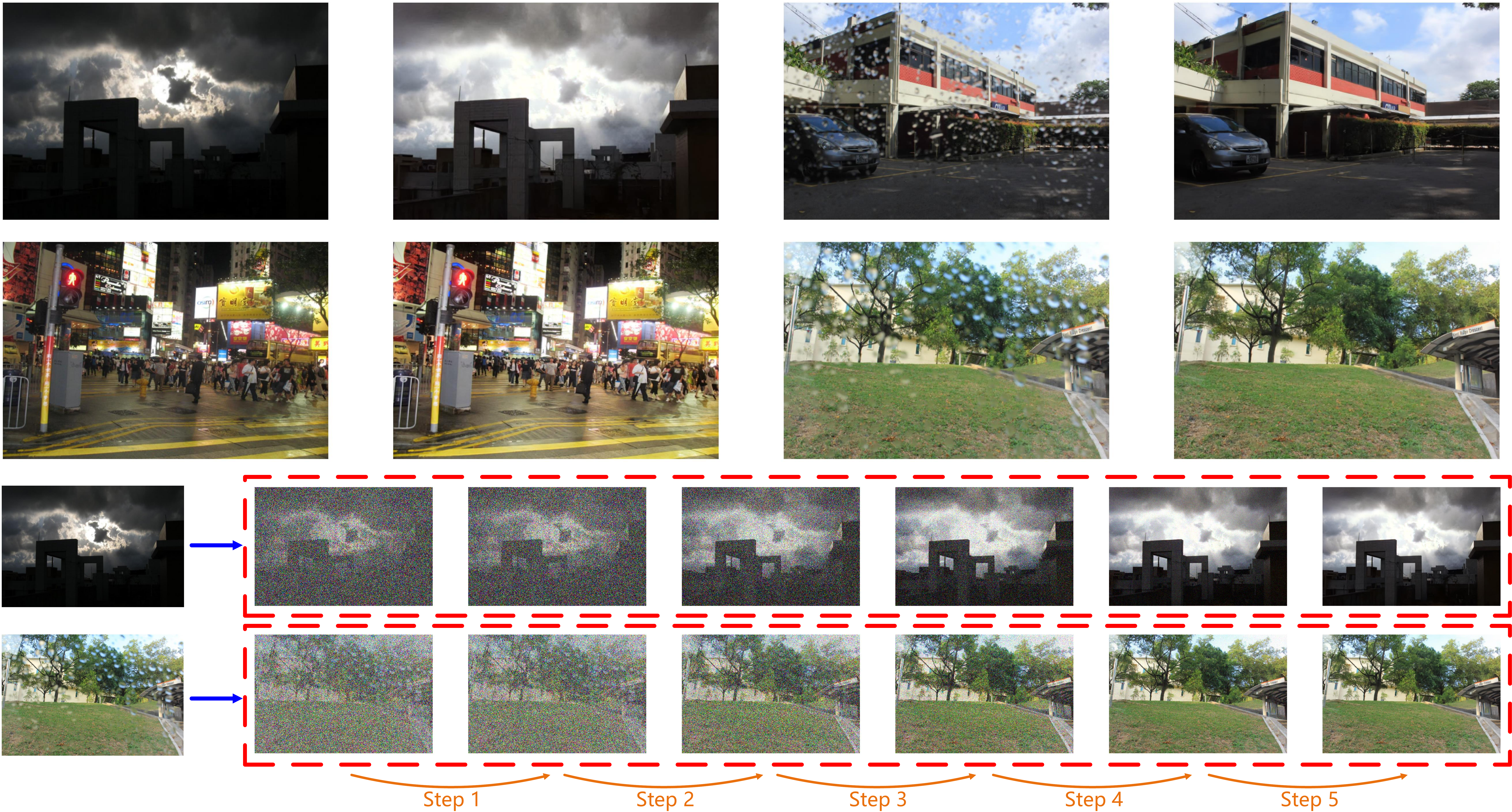}
    	\caption{Visualization of the five sampling steps, where the $\color{blue}{blue}$ arrow represents the smooth equivalence transformation, and the $\color{red}{red}$ box represents the resnoise-diffusion reverse process. We select the LOL web Test dataset (which \textbf{does not} have ground truth) and the Raindrop-B test dataset (which is much more challenging than Raindrop-A) to showcase the effects of low-light enhancement and deraining. We directly use the pretrained models on LOL dataset and Raindrop dataset, demonstrating the strong robustness of Resfusion.}
		\label{fig: visualization}
\end{figure*}
\section{Conclusion}
\label{sec: conclusion}
We propose the Resfusion, a general framework for image restoration. We explicitly provide the quantitative relationship between the residual term and the noise term, named as resnoise. Through employing the smooth equivalence transformation, we unify the training and inference process. Resfusion does not require any prior physical knowledge and can directly utilize existing noise schedules. Experimental results shows that Resfusion exhibits competitive performance for shadow removal, low-light enhancement, and deraining tasks, with only five sampling steps. It is important to note that Resfusion is not limited to image restoration and can be applied to any image generation domain. The versatility of our framework lies in its ability to simultaneously model the residual term and the noise term. Our subsequent experiments have demonstrated that Resfusion can be easily applied to various image generation tasks and exhibits strong competitiveness. 

\section{Acknowledgements}
\label{sec: acknowledgements}
This work is partially supported by the National Natural Science Foundation of China (62272248), the Natural Science Foundation of Tianjin (23JCZDJC01010, 23JCQNJC00010) and the Key Program of Science and Technology Foundation of Tianjin (24HHXCSS00004). 
\newpage

{
    \small
    \bibliographystyle{unsrtnat}
    \bibliography{reference}
}

\newpage

\appendix \section{Appendix Section}
\subsection{Detailed proof}
\label{sec: Detailed proof}
According to Eq.~\eqref{eq: x_t_resnoise} and Eq.~\eqref{eq: x_t_resnoise_reparameterized}, the forward process can be written as Eq.~\eqref{eq: q_sample_N_onestep} and Eq.~\eqref{eq: q_sample_N_fullstep}.
\begin{equation}
\label{eq: q_sample_N_onestep}
q(x_t|x_{t-1}, R)=\mathcal{N}(x_{t};\sqrt{\alpha_{t}} x_{t-1}+(1-\sqrt{\alpha_{t}})R,(1-\alpha_t)I)
\end{equation}
\begin{equation}
\label{eq: q_sample_N_fullstep}
q(x_t|x_0, R)=\mathcal{N}(x_{t};\sqrt{\overline\alpha_{t}} x_{0}+(1-\sqrt{\overline{\alpha}_{t}})R,(1-\overline{\alpha}_t)I)
\end{equation}
By simply performing a change of variable ($x_0 \rightarrow x_0-R$, $x_t \rightarrow x_t-R$, $x_{t-1} \rightarrow x_{t-1}-R$), the derivation of Eq.~\eqref{eq: q_sample_N} is identical in form to (71) - (84) in the reference~\cite{luo2022understanding}, where line 6 - 7 of Eq.~\eqref{eq: q_sample_N} corresponds to (73).
\begin{equation}
\label{eq: q_sample_N}
\begin{aligned}
& q(x_{t-1}|x_t, x_0, R) 
\\
& = \frac{q(x_{t}|x_{t-1}, x_{0}, R)q(x_{t-1}|x_0, R)}{q(x_{t}|x_{0}, R)}
\\
& = \frac{\mathcal{N}(x_{t};\sqrt{\alpha_{t}} x_{t-1}+(1-\sqrt{\alpha_{t}})R,(1-\alpha_t)I)\mathcal{N}(x_{t-1};\sqrt{\overline\alpha_{t-1}} x_{0}+(1-\sqrt{\overline{\alpha}_{t-1}})R,(1-\overline{\alpha}_{t-1})I)}{\mathcal{N}(x_{t};\sqrt{\overline\alpha_{t}} x_{0}+(1-\sqrt{\overline{\alpha}_{t}})R,(1-\overline{\alpha}_t)I)}
\\ 
& \propto exp\{-[\frac{[x_{t}-(\sqrt{\alpha_{t}} x_{t-1}+(1-\sqrt{\alpha_{t}})R)]^2}{2(1-\alpha_t)}+\frac{[x_{t-1}-(\sqrt{\overline\alpha_{t-1}} x_{0}+(1-\sqrt{\overline{\alpha}_{t-1}})R)]^2}{2(1-\overline{\alpha}_{t-1})} \\
&-\frac{[x_{t}-(\sqrt{\overline\alpha_{t}} x_{0}+(1-\sqrt{\overline{\alpha}_{t}})R)]^2}{2(1-\overline{\alpha}_t)}]\}
\\
& = exp\{-[\frac{[(x_{t}-R)-\sqrt{\alpha_{t}}(x_{t-1}-R)]^2}{2(1-\alpha_t)}+\frac{[(x_{t-1}-R)-\sqrt{\overline\alpha_{t-1}}(x_{0}-R)]^2}{2(1-\overline{\alpha}_{t-1})} \\
&-\frac{[(x_{t}-R)-\sqrt{\overline\alpha_{t}}(x_{0}-R)]^2}{2(1-\overline{\alpha}_t)}]\}
\\
& \propto \mathcal{N}(x_{t-1}-R;\frac{\sqrt{\alpha_{t}}(1-\overline{\alpha}_{t-1})(x_{t}-R)+\sqrt{\overline\alpha_{t-1}}(1-\alpha_{t})(x_{0}-R)}{1-\overline{\alpha}_{t}}, \widetilde\beta_{t}I) 
\\
& \propto \mathcal{N}(x_{t-1};\frac{\sqrt{\alpha_{t}}(1-\overline{\alpha}_{t-1})(x_{t}-R)+\sqrt{\overline\alpha_{t-1}}(1-\alpha_{t})(x_{0}-R)}{1-\overline{\alpha}_{t}}+R, \widetilde\beta_{t}I) 
\end{aligned}
\end{equation}
Then we can derive $\widetilde\mu(x_{t},x_{0}, R)$ as Eq.~\eqref{eq: widetilde_mu}.
\begin{equation}
\label{eq: widetilde_mu}
\widetilde\mu(x_{t},x_{0}, R) = \frac{\sqrt{\overline\alpha_{t-1}}\beta_{t}}{1-\overline{\alpha}_{t}}(x_{0}-R) + \frac{\sqrt{\alpha_{t}}(1-\overline{\alpha}_{t-1})}{1-\overline{\alpha}_{t}}(x_{t}-R) + R
\end{equation}
We can derive Eq.~\eqref{eq: x_t-R} through Eq.~\eqref{eq: x_t_resnoise_reparameterized}.
\begin{equation}
\label{eq: x_t-R}
x_{t} - R = \sqrt{\overline{\alpha}_{t}}(x_{0} - R) + \sqrt{1-\overline\alpha_{t}}\epsilon, \quad \epsilon\sim N(0,I)
\end{equation}
Thus we can derive Eq.~\eqref{eq: Lt-1} through Eq.~\eqref{eq: widetilde_mu} and Eq.~\eqref{eq: x_t-R}.
By simply performing a change of variables ($x_0 \rightarrow x_0-R$, $x_t \rightarrow x_t-R$), the derivation process becomes exactly identical in form to the derivation of equations (115) - (124) in the reference~\cite{luo2022understanding}, where Eq.~\eqref{eq: x_t-R} corresponds to (115) and Eq.~\eqref{eq: widetilde_mu} corresponds to (116).
\begin{small}
\begin{equation}
\label{eq: Lt-1}
\begin{aligned}
L_{t-1} - C 
& = \mathbb{E}_{x_{0},\epsilon,t}[\frac{1}{2\sigma_{t}^{2}}||\widetilde\mu(x_{t},x_{0}, R)-\mu_{\theta}(x_{t}(x_{0},\epsilon,R), t)||^2]
\\
& = \mathbb{E}_{x_{0},\epsilon,t}[\frac{1}{2\sigma_{t}^{2}}||\{\frac{1}{\sqrt{\alpha_{t}}}[(x_{t}(x_{0},\epsilon,R)-R)-\frac{\beta_{t}}{\sqrt{1-\overline{\alpha}_{t}}}\epsilon]+R\}-\mu_{\theta}(x_{t}(x_{0},\epsilon,R), t)||^2]
\end{aligned}
\end{equation}
\end{small}
According to Eq.~\eqref{eq: simplify}, we can modify Eq.~\eqref{eq: Lt-1} as Eq.~\eqref{eq: Lt-1_simple}.
\begin{small}
\begin{equation}
\label{eq: simplify}
\begin{aligned}
\frac{1}{\sqrt{\alpha_{t}}}(x_{t}-R-\frac{\beta_{t}}{\sqrt{1-\overline{\alpha}_{t}}}\epsilon)+R 
& = \frac{1}{\sqrt{\alpha_{t}}}(x_{t}-R+\sqrt{\alpha_{t}}R-\frac{\beta_{t}}{\sqrt{1-\overline{\alpha}_{t}}}\epsilon)
\\
& = \frac{1}{\sqrt{\alpha_{t}}}(x_{t}-(1-\sqrt{\alpha_{t}})R-\frac{\beta_{t}}{\sqrt{1-\overline{\alpha}_{t}}}\epsilon)
\\
& = \frac{1}{\sqrt{\alpha_{t}}}(x_{t}-\frac{(1-\sqrt{\alpha_{t}})\sqrt{1-\overline{\alpha}_{t}}}{\beta_{t}}\frac{\beta_{t}}{\sqrt{1-\overline{\alpha}_{t}}}R-\frac{\beta_{t}}{\sqrt{1-\overline{\alpha}_{t}}}\epsilon)
\\
& = \frac{1}{\sqrt{\alpha_{t}}}[x_{t}-\frac{\beta_{t}}{\sqrt{1-\overline{\alpha}_{t}}}(\epsilon+\frac{(1-\sqrt{\alpha_{t}})\sqrt{1-\overline{\alpha}_{t}}}{\beta_{t}}R)]
\end{aligned}
\end{equation}
\end{small}
\begin{small}
\begin{equation}
\label{eq: Lt-1_simple}
\begin{aligned}
&L_{t-1} - C  \\
&= \mathbb{E}_{x_{0},\epsilon,t}[\frac{1}{2\sigma_{t}^{2}}||\frac{1}{\sqrt{\alpha_{t}}}[x_{t}(x_{0},\epsilon,R)-\frac{\beta_{t}}{\sqrt{1-\overline{\alpha}_{t}}}(\epsilon+\frac{(1-\sqrt{\alpha_{t}})\sqrt{1-\overline{\alpha}_{t}}}{\beta_{t}}R)]-\mu_{\theta}(x_{t}(x_{0},\epsilon,R), t)||^2]
\end{aligned}
\end{equation}
\end{small}
According to \cite{ho2020denoising}, the minimize term become Eq.~\eqref{eq: resnoise}.

\subsection{Comparison with other methods}
\label{sec: comparison with other methods}
The main difference is in how the denoising diffusion, score, flow, or Schrödinger’s bridge are adapted to image restoration. Different methods select various elements as prediction targets: the noise term (Shadow Diffusion~\cite{guo2023shadowdiffusion}, SR3~\cite{saharia2022image}, WeatherDiffusion~\cite{ozdenizci2023restoring}), the residual term (DvSR~\cite{whang2022deblurring}, Rectified Flow~\cite{liu2022flow}), the target image (ColdDiffusion~\cite{bansal2024cold}, InDI~\cite{delbracio2023inversion}), or its linear transformation term (I$^2$SB~\cite{liu20232}). Similar to RDDM~\cite{liu2023residual}, Resfusion simultaneously predicts both the residual term and the noise term, and provides the quantitive relationship between them.

\textbf{Comparison with traditional diffusion-based methods}. Traditional diffusion-based image restoration methods~\cite{ozdenizci2023restoring, whang2022deblurring, wang2023lldiffusion, hou2024global, guo2023shadowdiffusion} adapt the diffusion model for image restoration tasks with degraded images as conditional input to implicitly guide the reverse generation process, without altering the original denoising diffusion process~\cite{ho2020denoising, song2020denoising}. Starting the reverse process from Gaussian white noise, traditional diffusion-based models consider only the degraded images as conditional input, resulting in an increased number of sampling steps. Meanwhile, these models are often task-specific, requiring the design of different model structures based on different scenarios. By introducing the residual term into the diffusion forward process, Resfusion bridge the gap between the input degraded images and ground truth, starting the reverse process directly from the noisy degraded images. As a versatile methodology for image restoration, Resfusion does not require any physical prior knowledge, and the image restoration can be completed in just five sampling steps.

\textbf{Comparison with RDDM~\cite{liu2023residual}}. RDDM can be seen as a diffusion process from the noisy input degraded image to the ground truth, while Resfusion represents a diffusion process from the noisy residual term to the ground truth. RDDM predicts the residual term and the noise term separately without specifying their weighted relationship, using a complex Automatic Objective Selection Algorithm (AOSA) to learn them. In contrast, Resfusion calculates the quantitative relationship between the residual term and the noise term, naming their weighted sum as resnoise. RDDM's forward process accumulates the residual term and the noise term, making its forward and backward processes inconsistent with DDPM, leading to poor generalization and interpretability. By transforming the learning of the noise term into the resnoise term, Resfusion's reverse inference process becomes consistent with DDPM, unifying the training and inference processes. Lastly, RDDM requires a customized noise schedule, as using existing noise schedules results in performance loss. Through the smooth equivalence transformation in resnoise-diffusion process, Resfusion can directly use the existing noise schedule.

\textbf{Comparison with Resshift~\cite{yue2023resshift}}. Similar to RDDM, Resshift's forward process also adopts an accumulation strategy for the residual term and the noise term. Therefore, Resshift also requires the design of a complex noise schedule, which is formulated as equation (10) in Resshift~\cite{yue2023resshift}. Resfusion can directly use the existing noise schedule instead of redesigning the noise schedule. The reverse process of Resshift is inconsistent with DDPM. The form of Resfusion's reverse inference process is consistent with the DDPM, leading to better generalization and interpretability. The prediction target of Resshift is $x_{0}$, while the prediction target of Resfusion is $res\epsilon$. Given that the essence of $res\epsilon$ is the noise term with an offset, and LDM models mainly predict the noise term, the loss function of Resfusion is extremely friendly to fine-tuning techniques such as Lora, which helps further scale up. Resshift diffuses in the latent space, utilizing the powerful encoding capability of models like VQ-GAN~\cite{esser2021taming}. Resfusion, on the other hand, directly diffuses in the RGB space. Resshift only explores fixed degradations such as image super-resolution. Resfusion explores more complex scenarios, including shadow removal, low-light enhancement, and deraining.

\textbf{Comparison with DvSR~\cite{whang2022deblurring}}. DvSR predicts clean images from input degraded images using a traditional (non-diffusion) network and calculates the residual term between the ground truth and the predicted clean images. DvSR employs denoising-based diffusion models to predict the noise term like DDPM, generating the residual term from Gaussian white noise. Unlike DvSR, Resfusion does not directly learn the residual term. Instead, it indirectly learns the distribution of the residual term through resnoise-diffusion process.

\textbf{Comparison with ColdDiffusion~\cite{bansal2024cold}}. ColdDiffusion aims to completely remove random noise from the diffusion model, replacing it with other transformations like blurring and masking. In contrast, Resfusion still incorporates noise diffusion. As shown in our ablation study, Resfusion requires the noise term for detail recovery. Since ColdDiffusion discards random noise, it needs additional degradation injection to enhance generation diversity. To simulate degradation processes for various restoration tasks, ColdDiffusion uses Gaussian blur for deblurring, snowification transform for snow removal, etc. These specific explorations might lose generality. Resfusion employs the residual term for directed diffusion from the ground truth to the noisy residual term, eliminating the need for task-specific degradation operators. Additionally, Resfusion provides solid theoretical derivation, whereas ColdDiffusion lacks a theoretical basis.

\subsection{Image translation}
\label{sec: image translation}
\begin{figure*}[h]
    \centering
    \includegraphics[width=\linewidth]{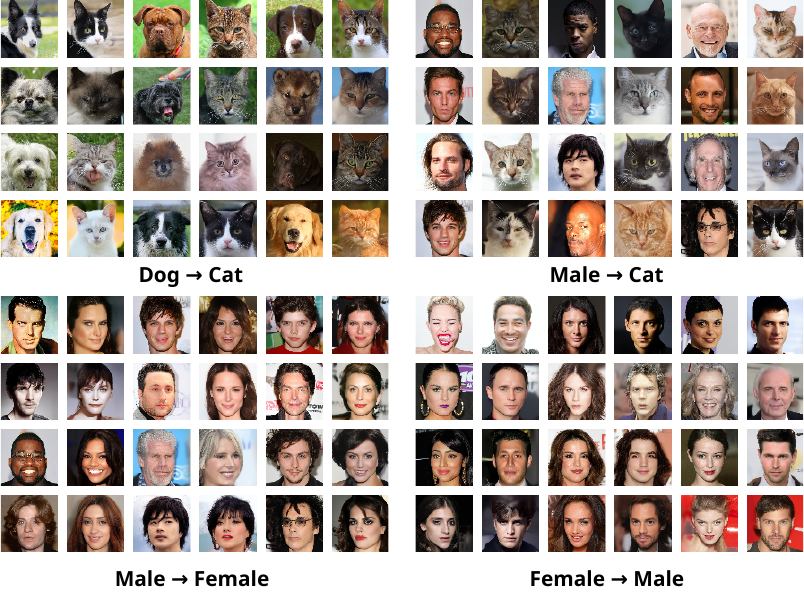}
    \caption{Visual results for image translation on the CelebA-HQ dataset and AFHQV2 dataset. The images are presented in pairs, with the translated image on the left and the target image on the right. We showcase the visual results of Resfusion for image translation tasks ``Dog $\rightarrow$ Cat'', ``Male $\rightarrow$ Cat'', ``Male $\rightarrow$ Female'', and ``Female $\rightarrow$ Male''.}
    \label{fig: translation}
\end{figure*}

Resfusion can also be implemented in image-to-image distribution transformation. By redefining $\hat{x}_{0}$ as the translated image and $x_{0}$ as the target image, we can easily transition Resfusion from image restoration to image translation. We train Resfusion with a truncated linear schedule on CelebA-HQ ($64\times64$) dataset~\cite{karras2017progressive} and AFHQV2 ($64\times64$) dataset~\cite{choi2020stargan} for image translation with 50 sampling steps. We selected the following image translation tasks: "Dog $\rightarrow$ Cat", "Male $\rightarrow$ Cat", "Male $\rightarrow$ Female", and "Female $\rightarrow$ Male". As shown in Fig.~\ref{fig: translation}, Resfusion can effectively model the shift between image domains, making it a unified methodology for a wider range of image generation tasks.

\subsection{Experimental setting details}
\label{sec: experimental setting details}
\begin{table*}[h]
\centering
\caption{Experimental settings for our Resfusion during the training stage.}
\quad
\scalebox{0.95}{
\begin{tabular}{c|c|c|c|c|c}
 \toprule
 \multirow{2}*{Tasks}     & \multicolumn{3}{c|}{Image Restoration} & Image               & Image                                                                                               \\
                          & Shadow Removal                         & Low-light           & Deraining                     & Generation              & Translation            \\
 \midrule
 \multirow{2}*{Datasets}  & \multirow{2}*{ISTD}                    & \multirow{2}*{LOL}  & \multirow{2}*{Raindrop}       & \multirow{2}*{CIFAR-10} & CelebA-HQ              \\
                          &                                        &                     &                               &                         & AFHQ-V2                 \\
 \hline
 Batch size               & 32                                     & 32                  & 32                            & 128                     & 128                    \\
 Image/patch size               & 256                                    & 256                 & 256                           & 32                      & 64                     \\
 $\hat{x}_{0}$            & $I_{in}$                               & $I_{in}$            & $I_{in}$                      & 0                       & $I_{in}$               \\
 Sampling steps           & 5                                      & 5                   & 5                             & 10 - 100                  & 50                     \\
 Learning rate            & 1.1e-4                                 & 1.1e-4              & 1.1e-4                        & 2e-4                    & 2e-4                   \\
 Training epochs          & 5k                                     & 5k                  & 5k                            & 3k                      & 3k                     \\
 \bottomrule
\end{tabular}
}
\label{tab: Experimental_settings}
\end{table*}
We use the PyTorch Lightning framework to train all the models, utilizing the AdamW~\cite{loshchilov2017decoupled} optimizer with the default settings of PyTorch~\cite{paszke2019pytorch}. Following \citet{liu2023residual}, we utilize the \href{https://github.com/Lyken17/pytorch-OpCounter}{THOP} to compute the number of parameters (Params) and multiply-accumulate operations (MACs). We only employ one U-net to predict resnoise and simply utilize a truncated Linear schedule across all tasks. For all the tasks, we implement a regular MSE loss as the loss function. The detailed experimental settings are provided in Table \ref{tab: Experimental_settings}. All experiments listed in Table \ref{tab: Experimental_settings} can be carried out with 8 NVIDIA RTX A6000 GPUs. 

\textbf{Image restoration}. We evaluate our method on several image restoration tasks, including shadow removal, low-light enhancement, and image deraining on 3 different datasets. For fair comparisons, the results of other image restoration methods are referenced from previously published papers~\cite{guo2023shadowformer,wang2023lldiffusion,wang2023ultra,ozdenizci2023restoring,liu2023residual} whenever possible. For all image restoration tasks, we used an identical U-net as the backbone, which is the same as RDDM~\cite{liu2023residual}. We take the shadow images and shadow masks together as the input condition (similar to \cite{liu2023residual, le2019shadow, zhu2022bijective}) for the ISTD dataset and only degraded images as the input condition for other datasets. We simply concatenate $x_t$ and $\hat{x}_{0}$ (and shadow masks) together in the channel dimension and feed them into the network. For the LOL dataset, we \textbf{do not} use pre-processing and post-processing techniques like Histogram Equalization and GT-mean. For the Raindrop dataset, we evaluate PSNR and SSIM based on the luminance channel Y of the YCbCr color space in accordance with previous work~\cite{liu2023residual, qian2018attentive, ozdenizci2023restoring}.
We employ a patch size of $256\times 256$ for all the datasets during the training stage. We use the peak signal-to-noise ratio (PSNR), structural similarity (SSIM), learned perceptual image patch similarity (LPIPS), and mean absolute error (MAE) as quantitative metrics.

\textbf{Image generation and image translation}. For image generation on the CIFAR10 ($32\times32$) dataset, we utilize the same U-net structure as DDIM~\cite{song2020denoising}. In contrast to DDIM which employs a linear schedule with $T=1000$ and a quadratic selection procedure to select sub-sampling steps, we use a linear schedule with $T=100$ and a linear selection procedure to select sub-sampling steps (for a fair comparison with truncated linear schedule) while training DDPM and DDIM. We use the Frechet Inception Distance (FID) as the quantitative metric. For image translation tasks, we employ a U-net structure with the same configuration as used in DDIM for CelebA~\cite{liu2015deep} as the backbone. To increase the diversity of the generated images, the translated images are not fed into the network as conditional input.

\subsection{Algorithm}
\label{sec: algrithm}
Based on the derivations from the Sec.~\ref{sec: learning the resnoise} and Sec.~\ref{sec: smooth equivalence transformation}, the training and inference processes of Resfusion can be represented as Algorithm \ref{alg: Training Algorithm for Resnoise Learning} and Algorithm \ref{alg: Inference Algorithm for Resfusion}. We highlight the modifications in our training and inference algorithms compared to DDPM in \textcolor{red}{red}. Just like the vanilla Denoising Diffusion Probabilistic Models (DDPM), Resfusion gradually fit $x_{t}$ to $x_{0}$, implicitly reducing the residual term between $\hat{x}_{0}$ and $x_{0}$ with the resnoise during the reverse inference process. Through transforming the learning of the noise term into the resnoise term, the form of resnoise-diffusion reverse inference process is consist with DDPM, leading to excellent interpretability.
\begin{figure*}[ht]
\center
\begin{minipage}[t]{0.47\textwidth}
\begin{algorithm}[H]
	\caption{Training Algorithm for Resfusion}
	\label{alg: Training Algorithm for Resnoise Learning}
	\begin{algorithmic}
    	\Require total diffusion steps $T$, degraded image and ground truth dataset $D={(\hat{x}_{0}^n, x_{0}^n)}_{n}^{N}$.
            \State $T'=\arg\min_{i=1}^{T} |{\sqrt{\overline\alpha_{i}}}-\frac{1}{2}|$
    	\Repeat
    	\State Sample $(\hat{x}_{0}^i, x_{0}^i)\sim D,\epsilon\sim N(0,I)$
    	\State Sample $t\sim Uniform({1,...,T'})$
            \State $\textcolor{red}{R = \hat{x}_{0} - x_{0}}$ 
            \State ${x}_{t}= \sqrt{\overline\alpha_{t}} x_{0} + 
            \textcolor{red}{(1-\sqrt{\overline{\alpha}_{t}})R} 
            + \sqrt{1-\overline\alpha_{t}}\epsilon$
            \State $res\epsilon = \epsilon + \textcolor{red}{\frac{(1-\sqrt{\alpha_{t}})\sqrt{1-\overline{\alpha}_{t}}}{\beta_{t}}R}$
            \State take gradient step on 
            \State \quad $\nabla_{\theta} ||\textcolor{red}{res\epsilon}-\textcolor{red}{res\epsilon_{\theta}}({x}_{t}, \hat{x}_{0},t)||^2$
    	\Until{convergence} 
	\end{algorithmic}
\end{algorithm}
\end{minipage}
\begin{minipage}[t]{0.52\textwidth}
\begin{algorithm}[H]
	\caption{Inference Algorithm for Resfusion}
	\label{alg: Inference Algorithm for Resfusion}
	\begin{algorithmic}
	\Require total diffusion steps $T$, degraded image $\hat{x}_{0}$, pretrained Resfusion model $res\epsilon_{\theta}$.
        \State $\widetilde\beta_{t}=\frac{1-\overline{\alpha}_{t-1}}{1-\overline{\alpha}_{t}}\beta_{t}$
        \State $\textcolor{red}{T'=\arg\min_{i=1}^{T} |{\sqrt{\overline\alpha_{i}}}-\frac{1}{2}|}$
	\State Sample $\epsilon\sim N(0,I)$
	\State $\textcolor{red}{x_{T'} = \sqrt{\overline\alpha_{T'}}\hat{x}_{0}+\sqrt{1-\overline\alpha_{T'}}\epsilon}$
	\For{$t=T', T'-1,...,2$}
	\State Sample $z\sim N(0,I)$
        \State $x_{t-1} =\frac{1}{\sqrt{\alpha_{t}}}(x_{t}-\frac{\beta_{t}}{\sqrt{1-\overline\alpha_{t}}}(\textcolor{red}{res\epsilon_{\theta}}(x_{t}, \hat{x}_{0},t)) + \sqrt{\widetilde{\beta}_{t}}z$
	\EndFor
        \State 
        \Return $x_{0}=\frac{1}{\sqrt{\alpha_{1}}}(x_{1}-\frac{\beta_{1}}{\sqrt{1-\overline\alpha_{1}}}\textcolor{red}{res\epsilon_{\theta}}(x_{1}, \hat{x}_{0},1))$
	\end{algorithmic}
\end{algorithm}
\end{minipage}
\end{figure*}

\subsection{Resource efficiency}
\label{sec: resource efficiency}
We compare the parameters, multiply-accumulate operation (MACs) and inference time with other image restoration methods on ISTD~\cite{wang2018stacked} dataset, LOL~\cite{wei2018deep} dataset and Raindrop~\cite{qian2018attentive} dataset by THOP, using $256\times256$ images as the input. For ISTD dataset, PSNR and SSIM are evaluated at a resolution of $256\times256$ after being resized. For LOL dataset and Raindrop dataset, the original image resolutions are maintained for the evaluation of PSNR and SSIM. The experimental results are quoted from the results of previous papers as well as our implementation based on open source code. 

As shown in Table \ref{tab: resource}, for the ISTD dataset, compared to Shadow Diffusion~\cite{guo2023shadowdiffusion}, Resfusion has 5$\times$ fewer parameters, 5$\times$ fewer sampling steps, and 20$\times$ fewer MACs. For the LOL dataset, compared to LLDiffusion~\cite{wang2023lldiffusion}, Resfusion has 6$\times$ fewer sampling steps. For the Raindrop dataset, compared to RainDiff$_{128}$~\cite{ozdenizci2023restoring}, Resfusion has 10$\times$ fewer parameters, 10$\times$ fewer sampling steps, and 50$\times$ fewer MACs. Experiments in shadow removal, low-light enhancement, and deraining demonstrate the effectiveness of Resfusion, enabling computationally constrained researchers to utilize our model for image restoration tasks.
\begin{table}[h]
\caption{Resource efficiency and performance analysis by THOP on ISTD dataset, LOL dataset and Raindrop dataset. ``MAC'' means multiply-accumulate operation. The best and second-best results are highlighted in \textbf{bold} and \underline{underlined}. ``$\uparrow$" (resp. ``$\downarrow$") means the larger (resp. smaller), the better. We use the symbol ``-" to indicate models or results that are unavailable.
\label{tab: resource}}
   \begin{center}
      \resizebox{\columnwidth}{!}{
         {
            \begin{tabular}{l|c|c|c|c|c}
               \toprule
               Methods & PSNR $\uparrow$ & SSIM $\uparrow$  & Params $\downarrow$ & MACs(G) $\times$ Steps $\downarrow$ & Inference Time (s) $\downarrow$\\
               \midrule
               \multicolumn{6}{c}{ISTD Dataset}\\
               \midrule
               Shadow Diffusion~\cite{guo2023shadowdiffusion} & \textbf{32.33} & \textbf{0.969} & \underline{55.5M} & \underline{182.1$\times$25 = 4552.5} & \underline{0.024$\times$25 = 0.600}\\
               SR3~\cite{saharia2022image} & 27.49 & 0.871 & 155.3M & 155.3$\times$100 = 15530.0 & -$\times$100 = -\\
               Resfusion (ours) & \underline{31.81} & \underline{0.965} & \textbf{7.7M} & \textbf{33.3$\times$5 = 167.5} & \textbf{0.027$\times$5 = 0.135}\\ 
               \midrule
               \multicolumn{6}{c}{LOL Dataset}\\
               \midrule
               LLFormer~\cite{wang2023ultra} & 23.65 & 0.816 & \underline{24.5M} & \textbf{22.0$\times$1 = 22.0} & \textbf{0.092$\times$1 = 0.092}\\
               LLDiffusion~\cite{wang2023lldiffusion} & \textbf{24.65} & \underline{0.843} &- &-$\times$30 = - & -$\times$30 = -\\
               Resfusion (ours) & \underline{24.63} & \textbf{0.860} & \textbf{7.7M} & \underline{32.9$\times$5 = 164.5} & \underline{0.027$\times$5 = 0.135}\\
               \midrule
               \multicolumn{6}{c}{Raindrop Dataset} \\
               \midrule
               RainDiff$_{64}$~\cite{ozdenizci2023restoring} & 32.29 & \textbf{0.942} & - & -$\times$10 = - & -$\times$10 = -\\
               RainDiff$_{128}$~\cite{ozdenizci2023restoring} & \underline{32.43} & 0.933 & 109.7M & 248.4$\times$50 = 12420.0 & -$\times$50 = -\\
               WeatherDiff$_{64}$~\cite{ozdenizci2023restoring} & 30.71 & 0.931 & \underline{82.9M} & \underline{463.1$\times$25 =  11577.5} & \underline{0.328$\times$25 = 8.20}\\
               WeatherDiff$_{128}$~\cite{ozdenizci2023restoring} & 29.66 & 0.923 & 85.6M & 261.8$\times$50 = 13090.0 & 0.439$\times$50 = 21.95\\
               Resfusion (ours) & \textbf{32.61} & \underline{0.938} & \textbf{7.7M} & \textbf{32.9$\times$5 = 164.5} & \textbf{0.027$\times$5 = 0.135}\\
               \bottomrule
            \end{tabular}
         }
    }
   \end{center}
\end{table}

\subsection{Truncated schedule}
\label{sec: Truncated schedule}
\begin{figure*}[h]
    \centering
    \includegraphics[width=\linewidth]{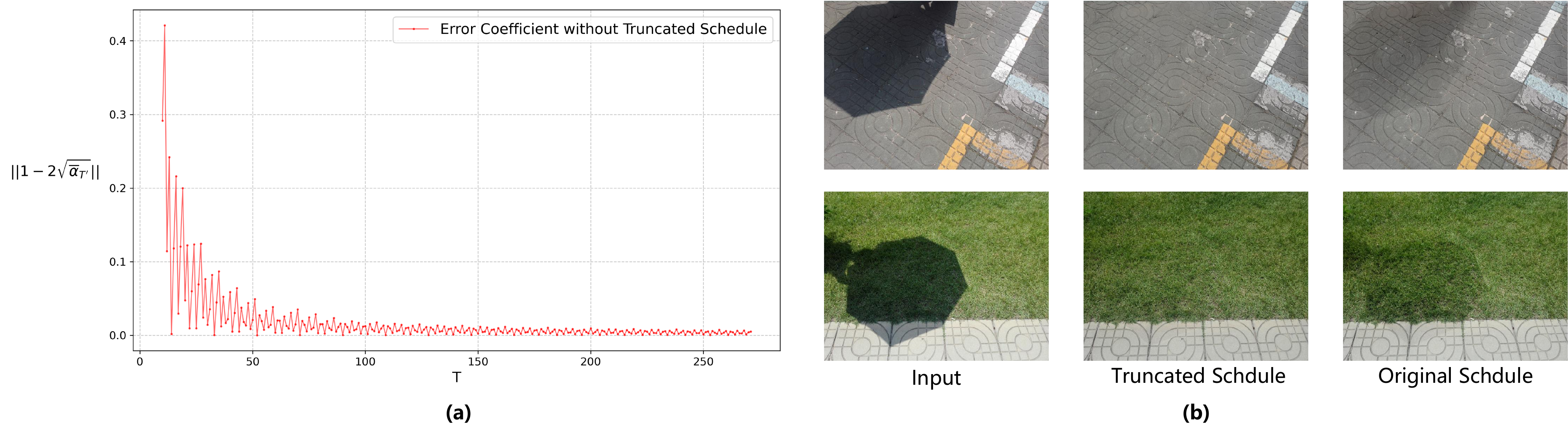}
    \caption{(a) Visualization of the relationship between the error coefficient and T. Technically, we can use the Truncated Schedule to eliminate this error when $T$ is small. (b) Visual comparisons between Truncated Schedule and Original Schedule under five sampling steps ($T'/T=5/12$). In terms of visual perception, the absence of Truncated Schedule will lead to residual shadows.}
    \label{fig: error analysis}
\end{figure*}
\begin{figure*}[h]
        \centering
	   \includegraphics[width=0.95\linewidth]{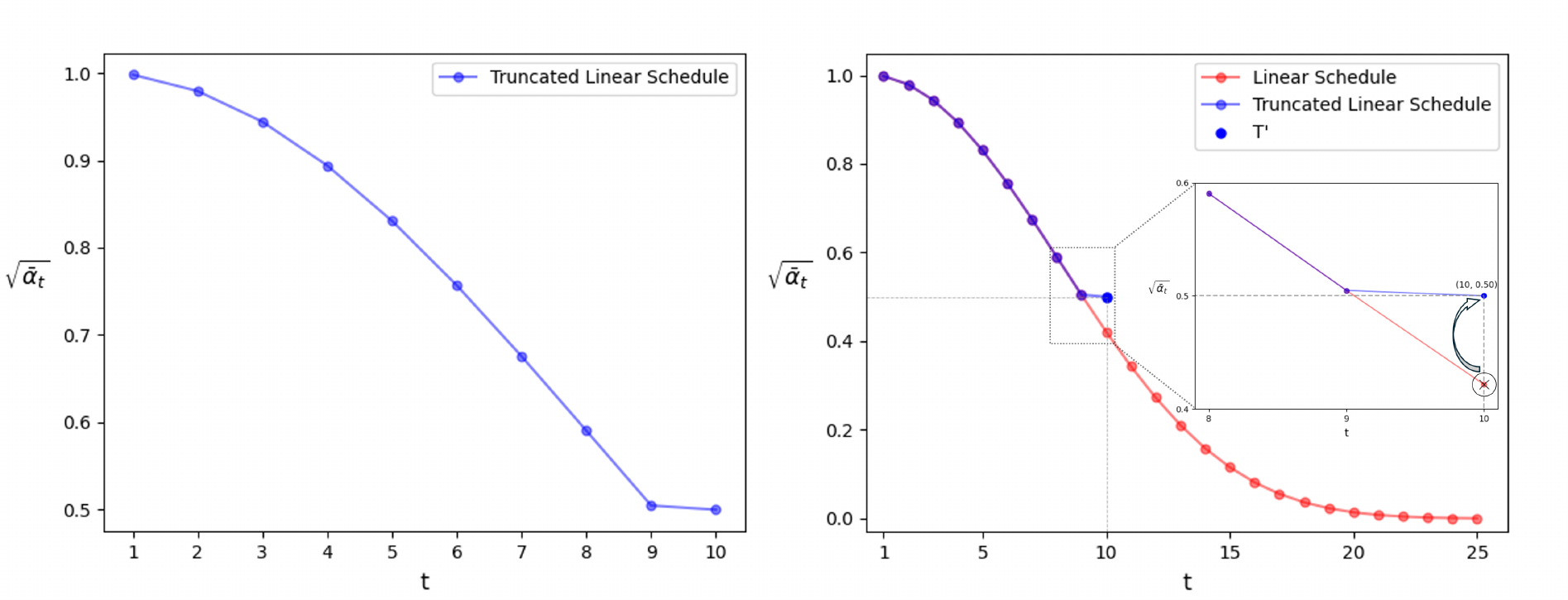}
	   \caption{The schematic diagram of our truncated linear schedule. Taking $T=25$, the left figure shows the 10 diffusion steps obtained by truncated linear schedule; the right figure shows the comparison of the truncated linear schedule and the linear schedule~\cite{nichol2021improved}.}
	   \label{fig: truncated}
\end{figure*}

We observed that in the actual diffusion forward process, the noise addition steps are uniformly spaced and discrete. The discontinuity of diffusion steps implies that when we approximate the acceleration point using Eq.~\eqref{eq: T'}, the offset of this approximate acceleration point relative to the ideal acceleration point is unavoidable, because ensuring the existence of intersection point with no offset requires that the $\color{gray}{gray}$ arrow and the $\color{violet}{violet}$ arrow in Fig.~\ref{fig: working principle} must be continuous. This offset actually quantifies the confidence level of the approximate equivalence $x_{T'} \approx \hat{x}_{T'}$ in Eq.~\eqref{eq: hat_P_x_t}. When T is small, the diffusion steps are divided sparsely, and the offset can be unacceptable. The absolute value of the offset can be derived as Eq.~\eqref{eq: absolute offset}.
\begin{equation}
\label{eq: absolute offset}
||x_{T'} - \hat{x}_{T'}||
= ||(2\sqrt{\overline\alpha_{T'}}-1) x_{0} + (1-2\sqrt{\overline\alpha_{T'}})\hat{x}_{0}||
= ||(1-2\sqrt{\overline\alpha_{T'}})R||
\end{equation}
As shown in Figure~\ref{fig: error analysis} (a), the absolute offset $||(1-2\sqrt{\overline\alpha_{T'}})R||$ exponentially decreases with the increase of $T$.
When $T$ is relatively small, this error is not negligible. However, this potential instability can be avoided in practical experiments, with a noise schedule named \textbf{Truncated Schedule} based on the existing noise schedules. In order to control the offset, we define an offset threshold $h$ with a default value of $0.01$. When decreasing $\sqrt{\overline\alpha}_{t}$, the first element less than $0.5$ is denoted as $\sqrt{\overline\alpha_{r}}$. If the difference between $0.5$ and $\sqrt{\overline\alpha_{r}}$ is greater than the offset threshold $h$, $\sqrt{\overline\alpha_{r}}$ will be reassigned to $0.5$ and the following elements will be truncated from here. Since the diffusion steps after the acceleration point is not involved in the actual diffusion process, direct truncation can avoid potential risks. Taking the truncated linear schedule~\cite{nichol2021improved} and $T=25$ as an example, Fig.~\ref{fig: truncated} demonstrates how to achieve the acceleration point $T'=10$. As shown in Figure~\ref{fig: error analysis} (b), Truncated Schedule can effectively eliminate "residual shadows".

\subsection{LOL-v2-real dataset}
\begin{figure*}[h]
    \centering
    \includegraphics[width=\linewidth]{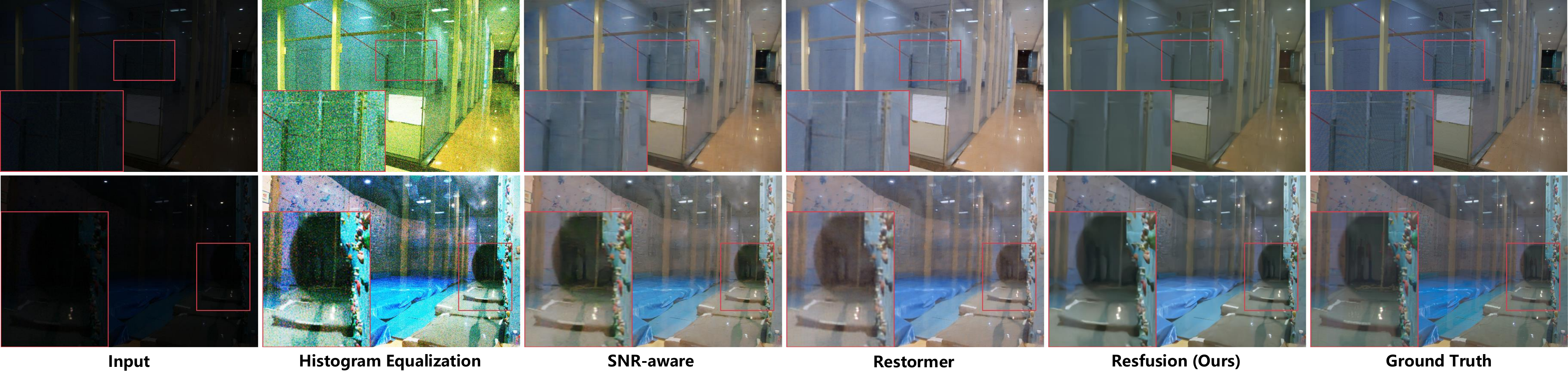}
    \caption{Visual comparisons of the restored results by different image restoration methods on the LOL-v2-real dataset.}
    \label{fig: lolv2}
\end{figure*}

\begin{table}[h]
    \centering
    \caption{Quantitative comparisons with other low-light enhancement methods on LOL-v2-real dataset. We report PSNR, SSIM and LPIPS. The best and second-best results are highlighted in \textbf{bold} and \underline{underlined}. ``$\uparrow$" (resp. ``$\downarrow$") means the larger (resp. smaller), the better.}%
    \begin{tabular}{l| ccc}
    \toprule
    \multicolumn{4}{c}{LOL-v2-real~\cite{yang2021sparse}}\\
    \midrule
    Method & PSNR $\uparrow$ & SSIM $\uparrow$ & LPIPS $\downarrow$\\
    \midrule
    Restormer~\cite{zamir2022restormer} & 18.69 & \underline{0.834} & 0.232\\
    LLFormer~\cite{wang2023ultra}  & \underline{20.06} & 0.792 & \underline{0.211}\\
    \textbf{Resfusion (ours)} & \textbf{22.06} & \textbf{0.839} & \textbf{0.175}\\
    \bottomrule
    \end{tabular}
    \label{tab: lolv2}
\end{table}
The LOL-v2-real dataset~\cite{yang2021sparse} includes visual degradations such as decreased visibility, intensive noise, and biased color. It contains 689 image pairs of both low-light and normal-light versions for training and 100 image pairs for evaluation. All experimental settings are exactly the same as the LOL dataset.
As shown in Figure~\ref{fig: lolv2}, compared to Histogram Equalization, 
Resfusion can significantly reduce noise, while also achieving a better color offset, demonstrating strong denoising capabilities.
We provide results in terms of PSNR, SSIM, and LPIPS on LOL-v2-real dataset in Table~\ref{tab: lolv2}.

\subsection{Limitations and Future work}
\label{sec: Limitations and Future work}
\textbf{Task specific}. Our main effort has been directed towards creating a general prototype model for image restoration and generation. This approach may lead to some performance limitations when compared to task-specific state-of-the-art methods~\cite{he2023reti, guo2023shadowdiffusion}. To enhance performance for particular tasks, potential strategies include employing task-specific backbones, incorporating physical prior knowledge, and utilizing customized noise schedules.

\textbf{Feature fusion}. In the reverse process, we simply concatenate the noisy image $x_{t}$ at time step $t$ with the conditional input $\hat{x}_{0}$ in the channel dimension. It is worth exploring more efficient feature fusion strategies, such as cross-attention, multi-stage, multi-scale, and multi-branch.

\textbf{Latent space}. The diffusion process in Resfusion is conducted in the original pixel space. Some studies~\cite{rombach2022high, luo2023refusion, yue2023resshift} have shown that conducting diffusion process in the latent space~\cite{esser2021taming, van2017neural} can significantly reduce computational complexity while ensuring the quality of generated images, which is worth exploring in the future.

\newpage

\subsection{More results and failure cases}
\label{sec: More results}

\begin{figure*}[h]
        \centering
	   \includegraphics[width=\linewidth]{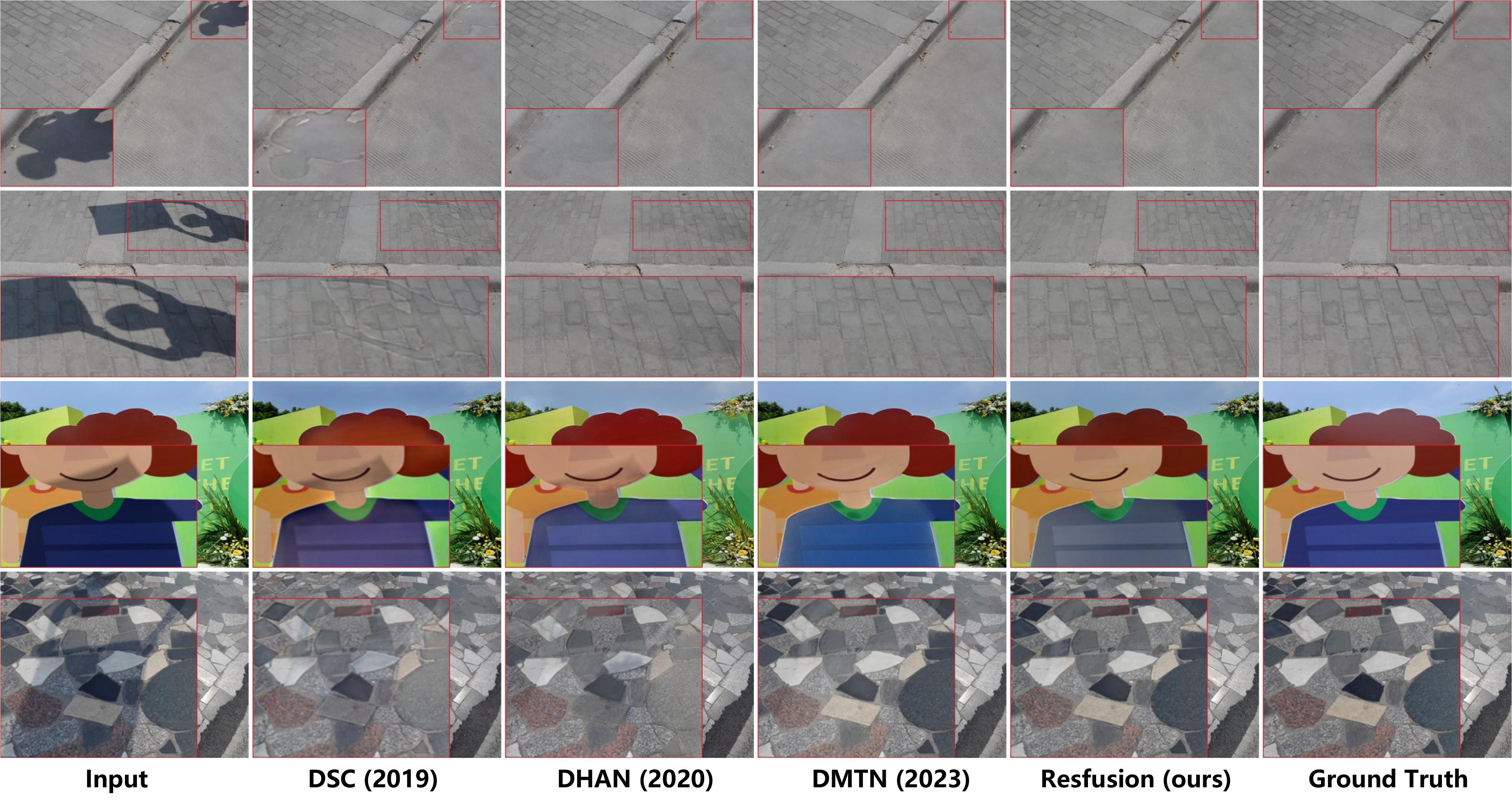}
	   \caption{More visual comparisons of the restored results by different shadow-removal methods on the ISTD dataset.}
	   \label{fig: istd4}
\end{figure*}

\begin{figure*}[h]
        \centering
	   \includegraphics[width=\linewidth]{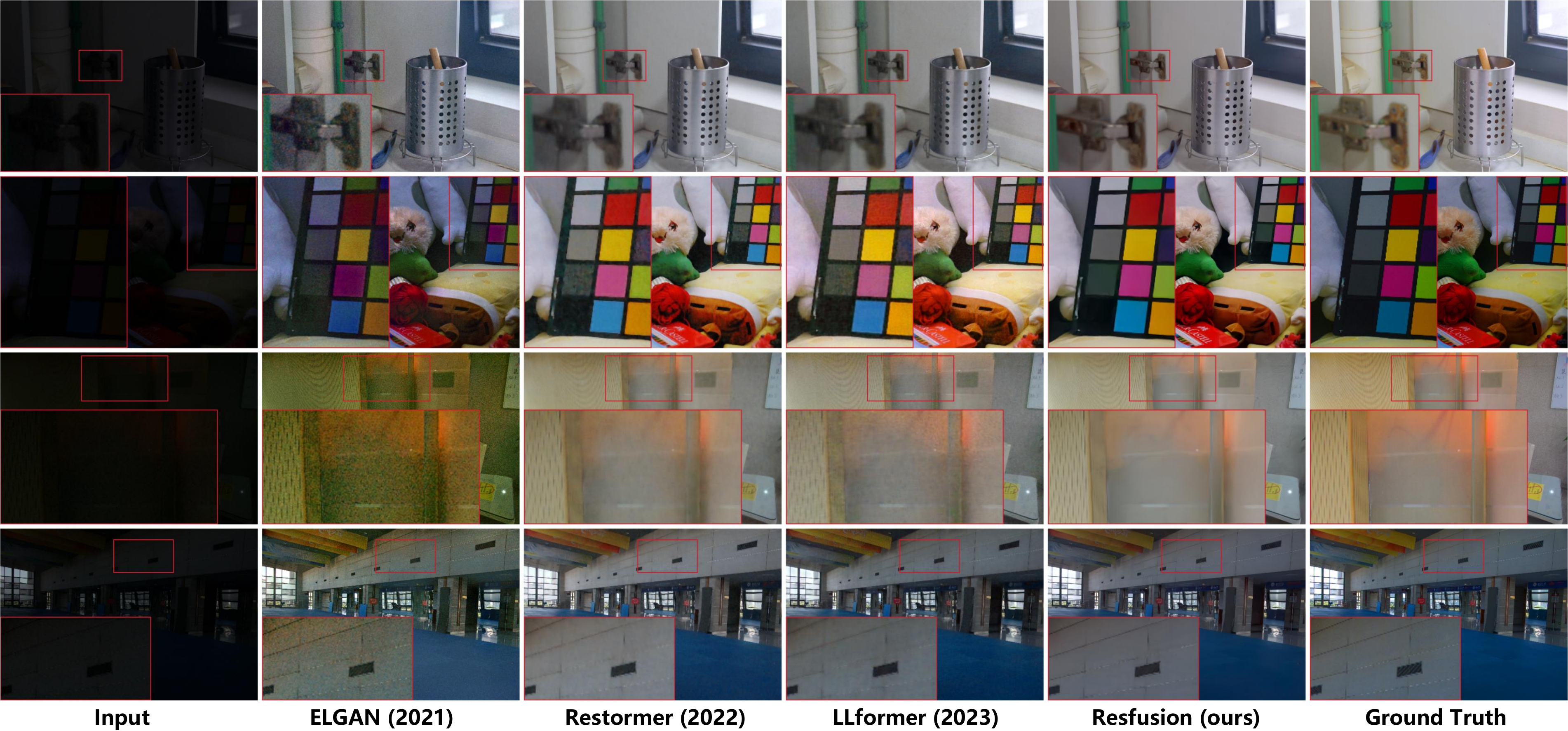}
	   \caption{More visual comparisons of the restored results by different low-light enhancement  methods on the LOL dataset.}
	   \label{fig: lol4}
\end{figure*}

\newpage

\begin{figure*}[h]
        \centering
	   \includegraphics[width=\linewidth]{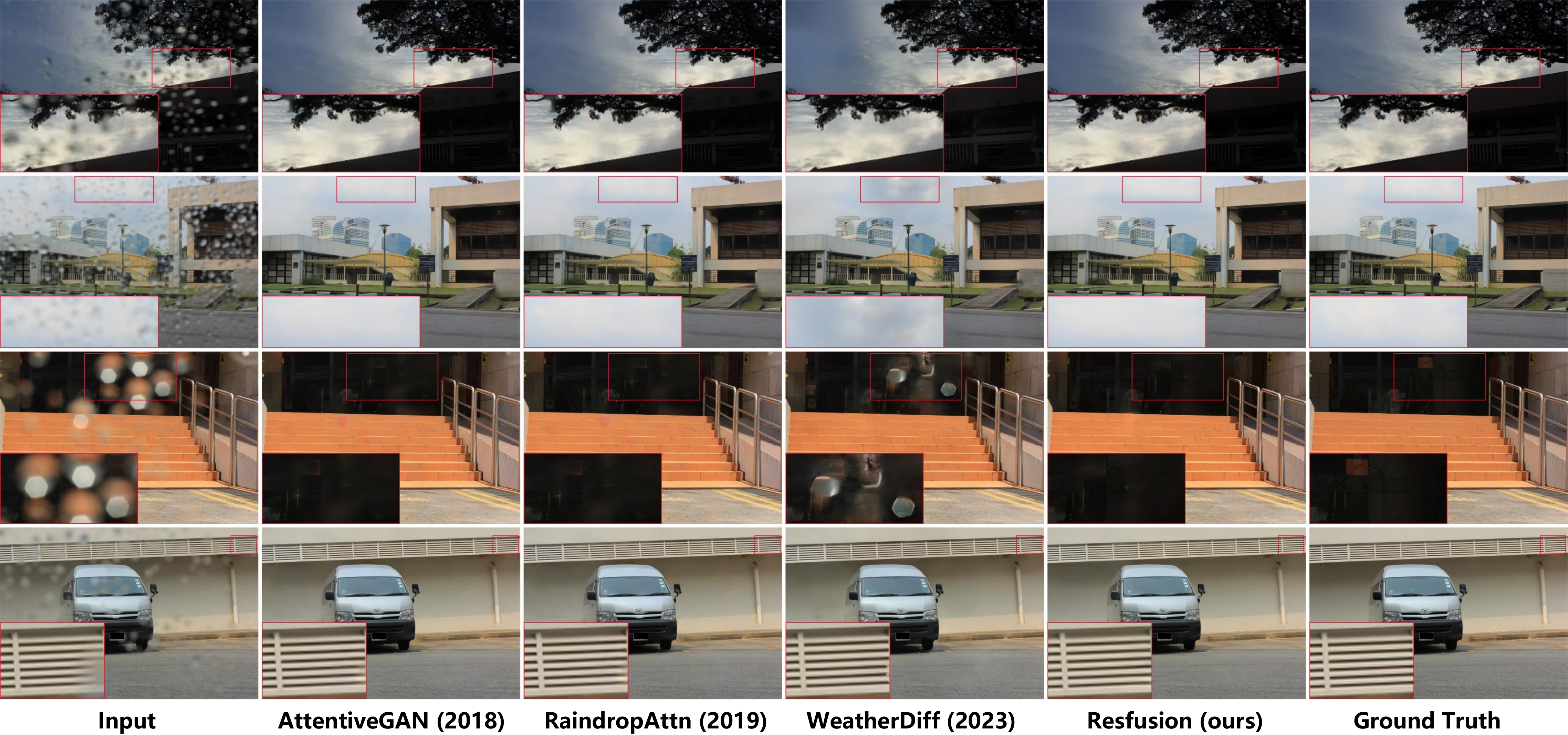}
	   \caption{More visual comparisons of the restored results by different deraining methods on the Raindrop dataset.}
	   \label{fig: Raindrop4}
\end{figure*}

\begin{figure*}[h]
        \centering
	   \includegraphics[width=\linewidth]{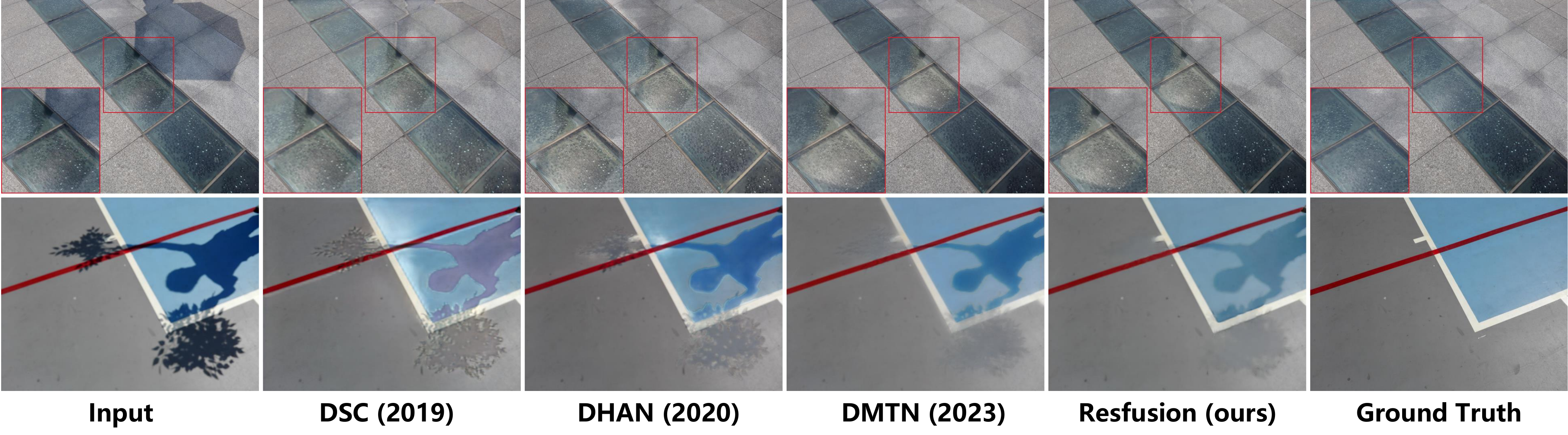}
	   \caption{Visual comparisons of the restored results by different shadow-removal methods on the ISTD dataset. (failure cases)}
	   \label{fig: fail_istd}
\end{figure*}

\begin{figure*}[h]
        \centering
	   \includegraphics[width=\linewidth]{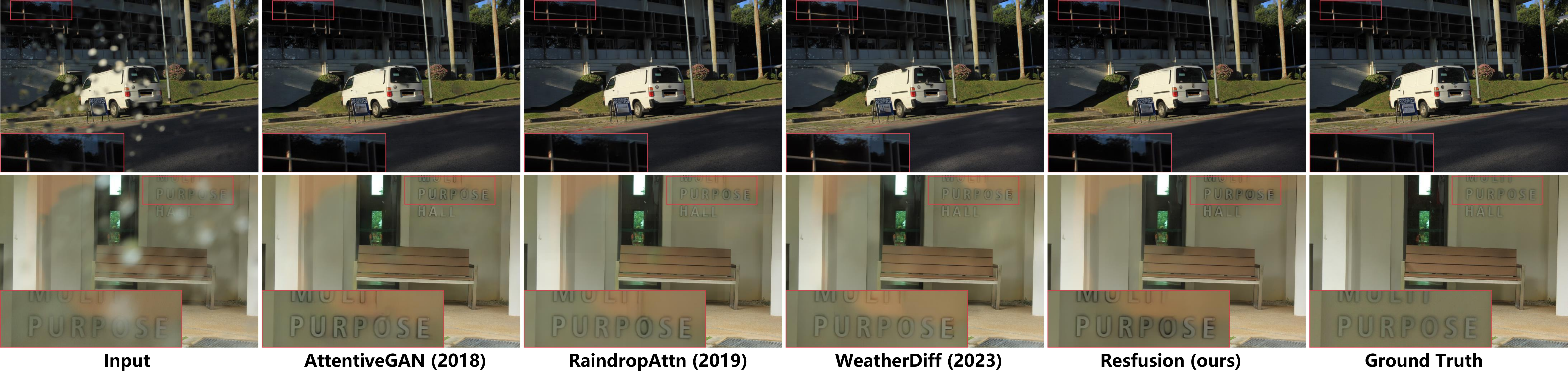}
	   \caption{Visual comparisons of the restored results by different deraining methods on the Raindrop dataset. (failure cases)}
	   \label{fig: fail_raindrop}
\end{figure*}

\newpage 

\end{document}